\documentclass[journal]{IEEEtran}
\usepackage{epsfig}

\usepackage{amsmath}
\usepackage{amssymb}
\usepackage{amsthm}
\usepackage{changes}
\usepackage{epstopdf}

\usepackage{algorithm}
\usepackage{algorithmic}
\usepackage{multirow}
\usepackage{color}
\usepackage{url}
\usepackage{cite}
\usepackage{wrapfig,lipsum,booktabs}
\usepackage{graphicx}
\usepackage{xcolor}
\usepackage[all]{xy}
\usepackage{flushend}
\usepackage{fancyhdr}

\usepackage{hyperref} 
 \hypersetup{
     bookmarks=true,         
     unicode=false,          
     pdftoolbar=true,        
     pdfmenubar=true,        
     pdffitwindow=false,     
     pdfstartview={FitH},    
     pdftitle={},    
     pdfauthor={},     
     pdfsubject={},   
     pdfcreator={},   
     pdfproducer={}, 
     pdfkeywords={Classification}{Cloud detection}{Random Fourier Features}{Variational inference}{Gaussian process}{kernels}, 
     pdfnewwindow=true,      
     colorlinks=true,       
     linkcolor={blue},          
     citecolor=blue,        
     filecolor=blue,      
     urlcolor=blue           
 }

\usepackage{setspace}

\usepackage{textcomp}
\usepackage{gensymb}

\newcommand{\R}{\mathbb{R}}

\DeclareMathOperator*{\diag}{diag}
\def\bv{{\mathbf{v}}}
\def\bw{{\mathbf{w}}}
\def\bx{{\mathbf{x}}}
\def\by{{\mathbf{y}}}
\def\bz{{\mathbf{z}}}
\def\b0{{\mathbf{0}}}

\def\bI{{\mathbf{I}}}

\def\bK{{\mathbf{K}}}

\def\bW{{\mathbf{W}}}
\def\bX{{\mathbf{X}}}

\def\bZ{{\mathbf{Z}}}

\def\bbeta{{\boldsymbol{\beta}}}

\def\bmu{{\boldsymbol{\mu}}}

\def\bxi{{\boldsymbol{\xi}}}

\def\bLambda{{\boldsymbol{\Lambda}}}

\def\bSigma{{\boldsymbol{\Sigma}}}

\newcommand{\md}{{\mathrm d}}
\newcommand{\p}{{\mathrm p}}

\usepackage{changes}
\definechangesauthor[name={Rafael Molina}, color=red]{rms}
\definechangesauthor[name={Pablo Morales}, color=brown]{pma}
\setremarkmarkup{(#2)}

\def\adp{\added[id=pma]}

\ifCLASSINFOpdf
\else
\fi
\begin{document}

%
\title{Remote Sensing Image Classification with\\ Large Scale Gaussian Processes}
%
%
%

\author{Pablo~Morales-\'{A}lvarez, 
Adri\'{a}n~P\'{e}rez-Suay,~\IEEEmembership{Member, IEEE}\\
Rafael~Molina, ~\IEEEmembership{Senior Member,~IEEE} 
and~Gustau~Camps-Valls,~\IEEEmembership{Senior Member,~IEEE}
\thanks{\copyright IEEE. Personal use of this material is permitted. Permission from IEEE must be obtained for all other users, including reprinting/republishing this material for advertising or promotional purposes, creating new collective works for resale or redistribution to servers or
lists, or reuse of any copyrighted components of this work in other works. DOI: 10.1109/TGRS.2017.2758922}
\thanks{Research funded by the European Research Council (ERC) under the ERC-CoG-2014 SEDAL project (grant agreement 647423), the Spanish Ministry of Economy and Competitiveness (MINECO) through projects TIN2013-43880-R, TIN2015-64210-R, and DPI2016-77869-C2-2-R, TEC2016-77741-R, the EUMETSAT through contract EUM/RSP/SOW/14/762293, and the Spanish Excellence Network TEC2016-81900-REDT. P. Morales-\'Alvarez also acknowledges financial support from \emph{La Caixa} Foundation.}
\thanks{P. Morales-\'Alvarez and R. Molina are with the Department of Computer Science and Artificial Intelligence, University of Granada, Spain (e-mail: \mbox{pablomorales@decsai.ugr.es}; rms@decsai.ugr.es).}
\thanks{A. P\'{e}rez-Suay and G. Camps-Valls are with the Image Processing Laboratory, Universitat de Val\'{e}ncia, Spain (e-mail: adrian.perez@uv.es; \mbox{gustau.camps@uv.es}).}
}

\maketitle
\begin{abstract}

Current remote sensing image classification problems have to deal with an unprecedented amount of heterogeneous and complex data sources. Upcoming missions will soon provide large data streams that will make land cover/use classification difficult. Machine learning classifiers can help at this, and many methods are currently available. A popular kernel classifier is the Gaussian process classifier (GPC), since it approaches the classification problem with a solid probabilistic treatment, thus yielding confidence intervals for the predictions as well as very competitive results to state-of-the-art neural networks and support vector machines. However, its computational cost is prohibitive for large scale applications, and constitutes the main obstacle precluding wide adoption. This paper tackles this problem by introducing two novel efficient methodologies for Gaussian Process (GP) classification. We first include the standard random Fourier features approximation into GPC, which largely decreases its computational cost and permits large scale remote sensing image classification. In addition, we propose a model which avoids randomly sampling a number of Fourier frequencies, and alternatively {\em learns} the optimal ones within a variational Bayes approach. The performance of the proposed methods is illustrated in complex problems of cloud detection from multispectral imagery and infrared sounding data.
Excellent empirical results support the proposal in both computational cost and accuracy.
\end{abstract}

\begin{IEEEkeywords}
Gaussian Process Classification (GPC),
random Fourier features, Variational Inference, Cloud detection, Seviri/MSG, IAVISA, IASI\slash AVHRR
\end{IEEEkeywords}

%
\IEEEpeerreviewmaketitle

\section{Introduction}\label{sec:intro}
%
%
%
%

\begin{flushright}
\small 
{\em ``... Nature almost surely operates by combining chance with necessity, randomness with determinism...''\\
--Eric Chaisson, Epic of Evolution: Seven Ages of the Cosmos }
\end{flushright}

\IEEEPARstart{E}{arth}-observation (EO) satellites provide a unique source of information to address some of the challenges of the Earth system science~\cite{Berger12}.
Current EO applications for image classification have to deal with a huge amount of heterogeneous and complex data sources. 

The super-spectral Copernicus Sentinels~\cite{Drusch2012,Donlon12}, as well as the planned EnMAP~\cite{Stuffler2007}, HyspIRI~\cite{Roberts2012}, PRISMA~\cite{Labate2009} and FLEX~\cite{Kraft2013}, will soon provide unprecedented data streams to be analyzed. Very high resolution (VHR) sensors like Quickbird, Worldview-2 and the recent Worldview-3~\cite{lpbc14} also pose big challenges to data processing. The challenge is not only attached to optical sensors. Infrared sounders, like the Infrared Atmospheric Sounding Interferometer (IASI)~\cite{TOURNIER2002} sensor on board the MetOp satellite series, impose even larger constraints: the orbital period of Metop satellites (101 minutes), the large spectral resolution (8461 spectral channels between 645~cm$^{-1}$ and 2760~cm$^{-1}$), and the spatial resolution (60$\times$1530 samples) of the IASI instrument yield several hundreds of gigabytes of data to be processed daily. The IASI mission delivers approximately $1.3\times10^6$ spectra per day, which gives a rate of about 29 Gbytes/day to be processed. EO radar images also increased in resolution, and current platforms such as ERS-1/2, ENVISAT, RadarSAT-1/2, TerraSAR-X, and Cosmo-SkyMED give raise to extremely fine resolution data that call for advanced scalable processing methods. Besides, we should not forget the availability of the extremely large remote sensing data archives\footnote{The Earth Observing System Data and Information System (EOSDIS) for example is managing around 4 terabytes daily, and the flow of data to users is about 20 terabytes daily.} already collected by several past missions. In addition, we should be also prepared for the near future in diversity and complementarity of sensors\footnote{Follow the links for an up-to-date list of current \href{https://earth.esa.int/web/guest/missions/esa-eo-missions}{ESA}, \href{http://www.eumetsat.int/website/home/Satellites/index.html}{EUMETSAT}, \href{http://global.jaxa.jp/projects/}{JAXA}, \href{http://www.cnsa.gov.cn/n6443408/index.html}{CNSA} and \href{https://www.nasa.gov/missions}{NASA} EO missions.}. These large scale data problems require enhanced processing techniques that should be accurate, robust and fast. Standard classification algorithms cannot cope with this new scenario efficiently.

In the last decade, kernel methods have dominated the field of remote sensing image classification~\cite{CampsValls09wiley,CampsValls11mc}. 
In particular, a kernel method called support vector machine ({SVM},~\cite{huang02,campsieee04,Mel04b,foody04,campsvallstgars05}) was gradually introduced in the field, and quickly became a standard for image classification. Further SVM developments considered the simultaneous integration of spatial, spectral and temporal information~\cite{Benediktsson05,fauvel08,Pac08,Tuia09tgrs,cam08}, the richness of hyperspectral imagery~\cite{campsvallstgars05,Plaza09}, and exploited the power of clusters of computers~\cite{Pla08b,Mun09b}. 
Undoubtedly, kernel methods have been the most widely studied classifiers, and became the preferred choice for users and practitioners. 
However, they are still not widely adopted in real practice because of the high computational cost when dealing with large scale problems.
Roughly speaking, given $n$ examples available for training, kernel machines need to store kernel matrices of size $n\times n$, and to process them using standard linear algebra tools (matrix inversion, factorization, eigen-decomposition, etc.) that typically scale cubically, ${\mathcal O}(n^3)$. This is an important constraint that hampers their applicability to large scale EO data processing. 

An alternative kernel classifier to SVM is the Gaussian Process classifier (GPC)~\cite{Rasmussen2006}. 
GPC has appealing theoretical properties, as it approaches the classification problem with a solid probabilistic treatment, and very good performance in practice.
The GPC method was originally introduced in the field of remote sensing in~\cite{Bazi10}, where very good capabilities for land cover classification from multi/hyperspectral imagery were illustrated. Since then, GPC has been widely used in practice and extended to many settings: hyperspectral image classification~\cite{Yang15}, semantic annotation of high-resolution remote sensing images~\cite{Chen13}, change detection problems with semisupervised GPC~\cite{Chen13b}, or classification of images with the help of user's intervention in active learning schemes~\cite{ruiz:2014,Kalantari16}. 
Unfortunately, like any other kernel method, its computational cost is very large.
This is probably the reason why GPC has not yet been widely adopted by the geoscience and remote sensing community in large scale classification scenarios, despite its powerful theoretical background and excellent performance in practice. 

In GP for classification we face two main problems. First, the non-conjugate observation model for classification (usually based on the sigmoid, probit, or Heaviside step functions) renders the calculation of the marginal distribution needed for inference impossible.
The involved integrals are not analytically tractable, so one has to resort to numerical methods or approximations~\cite{Rasmussen2006}.
One could rely on Markov Chain Monte Carlo (MCMC) methods, but they are computationally far too expensive.
By assuming a Gaussian approximation to the posterior of the latent variables, one can use the Laplace approximation (LA) and the (more accurate) expectation propagation (EP)~\cite{Minka01,Kuss05}.
The observation model can also be bounded, leading to the variational inference approach \cite{chen:2014} that we use in this paper. 
Once the non-conjugacy of the observation model has been solved, the second problem is the inversion of huge matrices, which yields the unbearable $\mathcal{O}(n^3)$ complexity.
Notice that this is the only difficulty that appears when GP is used for regression, where the observation model can be analytically integrated out.
This efficiency problem could be addressed with recent Sparse GP approximations based on inducing points and approximate inference \cite{damianou_deep2015}, but they come at the price of a huge number of parameters to estimate.

In this paper, we introduce two alternative pathways to perform large scale remote sensing image classification with GPC.
First, following the ideas in~\cite{rahimiRFF}, we approximate the squared exponential (SE) kernel matrix of GPC by a linear one based on projections over a reduced set of random Fourier features (RFF). This novel method is referred to as RFF-GPC. It allows us to work in the primal space of features, which significantly reduces the computational cost of large scale applications. In fact, a recent similar approach allows for using millions of examples in SVM-based land cover classification and regression problems \cite{Perez17rks}.
The solid theoretical ground and the good empirical RFF-GPC performance make it a very useful method to tackle large scale problems in Earth observation.
However, RFF-GPC can only approximate (theoretically and in practice) a predefined kernel (the SE in this work), and the approximation does not necessarily lead to a discriminative kernel. These shortcomings motivate our second methodological proposal: we introduce a novel approach to avoid {\em randomly sampling} a number of Fourier frequencies, and instead we propose {\em learning} the optimal ones within the variational approach. 
Therefore, Fourier frequencies are no longer \emph{randomly sampled} and \emph{fixed}, but {\em latent variables} estimated directly from data via variational inference.
We refer to this method as VFF-GPC (Variational Fourier Features). 
The performance of RFF-GPC and VFF-GPC is illustrated in large and medium size real-world remote sensing image classification problems: (1) classification of clouds over landmarks from a long time series of Seviri/MSG (Spinning Enhanced Visible and Infrared Imager, Meteosat Second Generation) remote sensing images, and (2) cloud detection using IASI and AVHRR (Advanced Very High Resolution Radiometer) infrared sounding data, respectively. Excellent empirical results support the proposed large scale methods in both accuracy and computational efficiency. In particular, the extraordinary performance of VFF-GPC in the medium size data set justifies its use not only as a large scale method, but also as a general-purpose and scalable classification tool capable of learning an appropriate discriminative kernel.

The remainder of the paper is organized as follows.
Section \ref{sec:theory} reviews the RFF approximation and introduces it into GPC, deriving RFF-GPC and the more sophisticated VFF-GPC.
Section \ref{sec:data} introduces the two real-world remote sensing data sets used for the experimental validation. 
Section \ref{sec:experiments} presents the experimental results comparing the two proposed methods and standard GPC in terms of accuracy and efficiency.
Section \ref{sec:conclusions} concludes the paper with some remarks and future outlook.

\section{Large Scale Gaussian Process Classification}\label{sec:theory}
Gaussian Processes (GP) \cite{rasmussen06} is a probabilistic state-of-the-art model for regression and classification tasks.
In the geostatistic community, GP for regression is usually referred to as \emph{kriging}.
For input-output data pairs $\{(\bx_i,y_i)\}_{i=1}^n$, a GP models the underlying dependence from a function-space perspective, i.e. introducing latent variables $\{f_i = f(\bx_i)\in\R\}_{i=1}^n$ that jointly follow a normal distribution $\mathcal{N}\left(\mathbf{0},\bK = (k(\bx_i,\bx_j))_{1\leq i,j \leq n}\right)$.
The kernel function $k$ encodes the sort of functions $f$ favored, and $\bK$ is the so-called kernel matrix.
The observation model of the output $y$ given the latent variable $f$ depends on the problem at hand.
In binary classification (i.e. when $y\in\{0,1\}$), the (non-conjugate) logistic observation model is widely used. It is given by the sigmoid function as $\p(y=1|f) = \psi(f) = (1+\exp(-f))^{-1}\in (0,1)$.

\subsection{Random Fourier Features}\label{sec:theory_RFF}

The main issue with large scale applications of GP is its $\mathcal{O}(n^3)$ cost at the training phase, which comes from the $n\times n$ kernel matrix inversion. 
The work \cite{rahimiRFF} presents a general methodology (based on Bochner's theorem \cite{rudin2011}) to approximate any positive-definite shift-invariant kernel $k$ by a linear one. This is achieved by explicitly projecting the original $d$-dimensional data $\bx$ onto $\mathcal{O}(D)$ random Fourier features $\bz(\bx)$, whose linear kernel $k_L$ approximates $k$.
This linearity will enable us to work in the primal space of features and substitute $n\times n$ matrix inversions by $\mathcal{O}(D)\times\mathcal{O}(D)$ ones, resulting in a total $\mathcal{O}(nD^2+D^3)$ computational cost. 
In large-scale applications, one can set a $D\ll n$, and thus the obtained $\mathcal{O}(nD^2)$ complexity represents an important benefit over the original $\mathcal{O}(n^3)$. Moreover, the complexity at test is also reduced from $\mathcal{O}(n^2)$ to $\mathcal{O}(D^2)$, even becoming independent on $n$. 

In this work we use the well-known SE (or Gaussian) kernel $k(\bx,\bx') = \gamma\cdot\exp(-||\bx-\bx'||^2/(2\sigma^2))$. Following the methodology in \cite{rahimiRFF}, this kernel can be linearly approximated as 
\begin{align}
k(\bx,\bx')\approx k_L(\bx,\bx') = \gamma\cdot\bz(\bx)^\intercal\bz(\bx'),\label{eq:kerapprox}
\end{align}
where 
\begin{multline}\label{feats_general}
\bz(\bx)^\intercal = D^{-1/2} \cdot\left(\cos\left(\bw_1^\intercal \bx\right),\sin\left(\bw_1^\intercal \bx\right),\dots \right.\\
\left. \dots,\cos\left(\bw_D^\intercal \bx\right),\sin\left(\bw_D^\intercal \bx\right) \right)\in\R^{2D},
\end{multline}
and the \emph{Fourier frequencies} $\bw_i$ must be sampled from a normal distribution $\mathcal{N}(\mathbf{0},\sigma^{-2}\bI)$. As explained in \cite[Claim 1]{rahimiRFF}, this approximation exponentially improves with the number $D$ of Fourier frequencies used (and also exponentially worsens with $d$, the original dimension of $\bx$). However, obviously, increasing $D$ in our methods will go at the cost of increasing the $\mathcal{O}(nD^2)$ and $\mathcal{O}(D^2)$ complexities.
Other kernels different from the SE one could be used, but that would imply sampling from a different distribution.

Our novel RFF-GPC method considers a standard Bayesian linear model over these new features $\bz$. Such a linear model corresponds to GP classification with the linear kernel $k_L$ \cite[Chapter 6]{BishopBook}. Since $k_L$ approximates the SE kernel $k$, our RFF-GPC constitutes an approximation to GP classification with SE kernel. However, RFF-GPC is well suited for large scale applications, as it works in the primal space of $\mathcal{O}(D)$ features and thus presents a $\mathcal{O}(nD^2)$ (resp. $\mathcal{O}(D^2)$) train (resp. test) complexity.

Notice that RFF-GPC needs to sample the Fourier frequencies $\bw_i$ from $\mathcal{N}(\mathbf{0},\sigma^{-2}\bI)$ from the beginning, whereas hyperparameters $\sigma$ and $\gamma$ must be estimated during the learning process (just as in standard GP classification). In order to uncouple $\bw_i$ and $\sigma$, in the sequel we consider the equivalent features 
\begin{multline}\label{feats_RFF}
\bz(\bx|\sigma,\bW)^\intercal = D^{-1/2} \cdot\left(\cos\left(\sigma^{-1}\bw_1^\intercal \bx\right),\sin\left(\sigma^{-1}\bw_1^\intercal \bx\right),\dots \right.\\
\left. \dots,\cos\left(\sigma^{-1}\bw_D^\intercal \bx\right),\sin\left(\sigma^{-1}\bw_D^\intercal \bx\right) \right),
\end{multline}
with $\bw_i$ now sampled from $\mathcal{N}(\mathbf{0},\bI)$ and fixed. Notice that we have collectively denoted $\bW=[\bw_1,\dots,\bw_D]^\intercal\in\R^{D\times d}$.

At this point, it is natural to consider other possibilities for the Fourier frequencies $\bW$, rather than just randomly sample and fix them from the beginning. The proposed VFF-GPC model treats them as hyperparameters to be estimated (so as to maximize the likelihood of the observed data), just like $\sigma$ and $\gamma$ in the case of RFF-GPC. This makes VFF-GPC more expressive, flexible, and tailored to the data, although it may no longer constitute an approximation to the SE kernel (for which the $\bw_i$ must be normally distributed). More specifically, VFF-GPC does start with an approximated SE kernel (since the $\bw_i$ are initialized with a normal distribution), but the maximum a posteriori (MAP) optimization on $\bw_i$ makes it \emph{learn} a new kernel which may no longer approximate a SE one.
Therefore, for VFF-GPC we also use $\bz(\bx|\sigma,\bW)$ as in eq.~\eqref{feats_RFF}, with now both $\sigma$ and $\bW$ to be estimated.
Interestingly, VFF-GPC extends the Sparse Spectrum Gaussian Process model originally introduced for regression in \cite{SSGPR}, to GP classification. 

More specifically, the authors there also sparsify the SE kernel by working on the primal space of $\cos$ and $\sin$ Fourier features, see \cite[Equation 5]{SSGPR}. However, our classification setting involves a sigmoid-based (logistic) observation model for the output given the latent variable (see the next eq.~\eqref{obsModel}), whereas in regression this is just given by a normal distribution. Therefore, VFF-GPC needs to additionally deal with the non-conjugacy of the sigmoid, which motivates the variational bound of eq.~\eqref{varbound} and the consequent variational inference procedure described in Section \ref{sec:varInf}. It is interesting to realize that VFF-GPC is introduced here as a natural extension of RFF-GPC, whereas there is not a regression analogous for RFF-GPC in \cite{SSGPR}.

Another possibility for the Fourier frequencies would be to estimate them (just as in VFF-GPC) but considering alternative prior distributions $\p(\bW)$ (which means utilizing alternative kernels).
Moreover, instead of maximum a posteriori inference, we could address the marginalization of the Fourier frequencies $\bW$.
Alternatively, to promote sparsity, the use of Gaussian Scale Models (GSM) \cite{babacan:2012} could also be investigated.
These possibilities will be explored in future work, and here we will concentrate on RFF-GPC and VFF-GPC.

\subsection{Models formulation}

As anticipated in previous section, RFF-GPC and VFF-GPC are standard Bayesian linear models working on the explicitly mapped features $\bz(\bx|\sigma,\bW)$ of eq.~\eqref{feats_RFF}. In the case of RFF-GPC, $\bW$ is sampled from $\mathcal{N}(\mathbf{0},\bI)$ at the beginning and fixed, with $\sigma$ to be estimated. In the case of VFF-GPC, both $\bW$ and $\sigma$ are estimated, with a $\mathcal{N}(\mathbf{0},\bI)$ prior over $\bW$. In order to derive both methods in a unified way, $\Phi$ will denote $\sigma$ for RFF-GPC and both $(\bW,\sigma)$ for VFF-GPC.

Since we are dealing with binary classification, we consider the standard logistic observation model
\begin{equation}\label{obsModel}
    \p(y = 1 | \bbeta,\Phi,\bx) = \psi(\bbeta^\intercal\bz) =(1+\exp(-\bbeta^\intercal \bz))^{-1},
\end{equation}
where $\bz=\bz(\bx|\Phi)$. For the weights $\bbeta\in\R^{2D}$ we utilize the prior normal distribution $\p(\bbeta|\gamma) = \mathcal{N}(\bbeta|\mathbf{0},\gamma\bI)$, with $\gamma$ to be estimated, see eq.~(\ref{eq:kerapprox}). 

For an observed dataset $\mathcal{D}=\{(\bx_i,y_i)\}_{i=1}^n\subset\R^d\times\{0,1\}$, the joint p.d.f. reads
\begin{align}
\p(\by,\bbeta|\Phi,\gamma,\bX)&=\p(\by|\bbeta,\Phi,\bX)\p(\bbeta|\gamma) \nonumber \\
&= \left(\prod_{i=1}^n \p(y_i| \bbeta,\Phi,\bx_i)\right) \p(\bbeta|\gamma),\label{eq:jpdf}
\end{align}
where we collectively denote $\by = (y_1,\dots,y_n)^\intercal$ and $\bX = [\bx_1,\dots,\bx_n]^\intercal$. For the sake of brevity, from now on we will systematically omit the conditioning on $\bX$.

\subsection{Variational inference}\label{sec:varInf}

Given the observed dataset $\mathcal{D}$, in this section we seek point estimates of $\gamma$ and $\Phi$ by maximizing the marginal likelihood $\p(\by|\Phi,\gamma)$ (in VFF-GPC, the additional prior $\p(\bW)=\mathcal{N}(\bW|\mathbf{0},\bI)$ yields maximum a posteriori inference, instead of maximum likelihood one, for $\bW$).
After that, we obtain (an approximation to) the posterior distribution $\p(\bbeta|\by,\Phi,\gamma)$. Due to the non-conjugate observation model, the required integrals will be mathematically intractable, and we will resort to the variational inference approximation~\cite[Section 10.6]{BishopBook}.

First, notice that integrating out $\bbeta$ in eq.~\eqref{eq:jpdf} is not analytically possible due to the sigmoid functions in the observation model $\p(\by|\bbeta,\Phi)$. To overcome this problem, we use the variational bound
\begin{equation}\label{varbound}
\log\left(1+e^{x}\right) \leq \lambda(\xi)(x^2-\xi^2)+\frac{x-\xi}{2}+\log\left(1+e^{\xi}\right),
\end{equation}
which is true for any real numbers $x,\xi$ and where $\lambda(\xi) = (1/2\xi)\left( \psi(\xi)-1/2 \right)$ \cite[Section 10.6]{BishopBook}. Applying it to every factor of $\p(\by|\bbeta,\Phi)$, we have the following lower bound
\begin{equation}\label{varboundHere}
\p(\by|\bbeta,\Phi) \geq \exp\left(-\bbeta^\intercal \bZ^\intercal \bLambda\bZ \bbeta + \bv^\intercal\bZ\bbeta 
\right)\cdot C(\bxi).
\end{equation}
Here we write $\bZ=[\bz_1,\dots,\bz_n]^\intercal\in\R^{n\times 2D}$ for the projected-data matrix (which depends on $\Phi$), $\bLambda$ is the diagonal matrix $\diag(\lambda(\xi),\dots,\lambda(\xi_n))$, $\bv=\by-(1/2)\cdot\mathbf{1}_{n\times 1}$, and the term $C(\bxi) = \prod_i\exp\left(\lambda(\xi_i)\xi_i^2+(1/2)\xi_i-\log(1+\exp(\xi_i))\right)$ only depends on $\bxi$. The key is that this lower bound for $\p(\by|\bbeta,\Phi)$ is conjugate with the normal prior $\p(\bbeta|\gamma)$ (since it is the exponential of a quadratic function on $\bbeta$), and thus it allows us integrating out $\bbeta$ in eq~\eqref{eq:jpdf}. In exchange, we have introduced $n$ additional hyperparameters $\bxi=(\xi_1,\dots,\xi_n)^\intercal$ that will need to be estimated along with $\Phi$ and $\gamma$. 

Therefore, substituting $\p(\by|\bbeta,\Phi)$ for its bound, eq.~\eqref{eq:jpdf} can be lower bounded as
\begin{gather}\label{varGlobal}
\p(\by,\bbeta|\Phi,\gamma) \geq F(\by,\Phi,\gamma,\bxi)  \cdot\mathcal{N}(\bbeta|\bmu,\bSigma),
\end{gather}
where we have denoted
\begin{gather}
\nonumber F(\by,\Phi,\gamma,\bxi) = C(\bxi)\left\vert \gamma^{-1}\bSigma \right\vert^{1/2}\exp\left(\frac{1}{2}\bmu^\intercal\bSigma^{-1}\bmu\right), \\
\bSigma = \left(\bZ^\intercal(2\bLambda)\bZ+\gamma^{-1}\bI \right)^{-1}, \quad \bmu = \bSigma\bZ^\intercal\bv.\label{posterior}
\end{gather}

Now it is clear that we can marginalize out $\bbeta$, and iteratively estimate the optimal values of $\Phi$, $\gamma$ and $\bxi$ as those that maximize $F(\by,\Phi,\gamma,\bxi)$ (or equivalently $\log F(\by,\Phi,\gamma,\bxi)$) for the observed $\by$.
Starting at $\bxi^{(1)}$, $\Phi^{(1)}$, and $\gamma^{(1)}$, we can calculate  $\bxi^{(k)}$, $\Phi^{(k)}$, and $\gamma^{(k)}$ for $k \ge 1$.
In the case of $\bxi$, we make use of the local maximum condition $\partial F/\partial\bxi = 0$.
From there, it is not difficult to prove that the optimal value satisfies \cite{BishopBook}
\begin{equation}\label{updateXi}
    \bxi^{(k+1)} = \sqrt{\diag\left(\bZ^{(k)}\bSigma^{(k)}\left(\bZ^{(k)}\right)^\intercal\right)+\left(\bZ^{(k)}\bmu^{(k)}\right)^2},
\end{equation}
where the square and square root of a vector are understood as element-wise.
In the case of $\Phi$ and $\gamma$, we use nonlinear conjugate gradient methods \cite{minimize} and obtain (notice that for VFF-GPC we can collapse $\bW$ and $\sigma$, removing the prior on $\bW$)
\begin{align}\label{eq:Phi_gamma}
&\left(\Phi^{(k+1)},\gamma^{(k+1)}\right) = \arg\max_{\Phi,\gamma}\left\{ -\log\left\vert 2\gamma\bZ^\intercal\bLambda^{(k+1)}\bZ+\bI \right\vert  \right. \nonumber \\ &\hspace*{1cm}+ \left. \bv^\intercal \bZ \left( 2\bZ^\intercal\bLambda^{(k+1)}\bZ+\gamma^{-1}\bI\right)^{-1} \bZ^\intercal \bv \right\}.
\end{align}

Once the hyperparameters $\Phi$, $\gamma$, and $\bxi$ have been estimated by $\hat\Phi$, $\hat\gamma$, and $\hat\bxi$ respectively, we need to compute the posterior $\p(\bbeta|\by,\hat\Phi,\hat\gamma)$. As before, this is mathematically intractable due to the sigmoids in the observation model. Therefore, we again resort to the variational bound in eq.~\eqref{varGlobal} to get an optimal approximation $\hat\p(\bbeta)$ to the posterior $\p(\bbeta|\by,\hat\Phi,\hat\gamma)$. Namely, we do it by minimizing (an upper bound of) the KL divergence between both distributions:
\begin{align*}
& \mathrm{KL}\left(\hat\p(\bbeta)||\p(\bbeta|\by,\hat\Phi,\hat\gamma)\right) = \int\hat\p(\bbeta)\log\frac{\hat\p(\bbeta)}{\p(\bbeta|\by,\hat\Phi,\hat\gamma)}\md \bbeta   \\
&= \log\p(\by|\hat\Phi,\hat\gamma)+ \int\hat\p(\bbeta)\log\frac{\hat\p(\bbeta)}{\p(\by,\bbeta|\hat\Phi,\hat\gamma)}\md \bbeta  \\
&\leq \log\p(\by|\hat\Phi,\hat\gamma) - \log F(\by,\hat\Phi,\hat\gamma,\hat\bxi) + \mathrm{KL}\left(\hat\p(\bbeta)||\mathcal{N}(\bbeta|\hat\bmu,\hat\bSigma)\right).
\end{align*}
Thus, the minimum is reached for $\hat\p(\bbeta) = \mathcal{N}(\bbeta|\hat\bmu,\hat\bSigma)$, with $\hat\bmu$ and $\hat\bSigma$ calculated in  eq.~\eqref{posterior} using $\hat\Phi$, $\hat\gamma$, and $\hat\bxi$.

In summary, at training time, our methods RFF-GPC and VFF-GPC run iteratively until convergence of the hyperparameters $\Phi$, $\gamma$, and $\bxi$ to their optimal values $\hat \Phi$, $\hat\gamma$, and $\hat\bxi$ (see Algorithm \ref{algorithm}). The computations involved there suppose a computational complexity of $\mathcal{O}(nD^2+D^3)$ (which equals $\mathcal{O}(nD^2)$ when $n\gg D$), whereas standard GPC scales as $\mathcal{O}(n^3)$. 
At test time, the probability of class $1$ for a previously unseen instance $\bx_*\in\R^d$ is:
\begin{align}\label{eq:test}
\p(y_*=1) &\approx
\int \p(y = 1|\bbeta,\hat \Phi,\bx_*)\hat\p(\bbeta){\rm d}\bbeta \approx \nonumber \\
& \approx \psi\left(\hat\bz_*^\intercal \hat\bmu\cdot\left(1+(\pi/8)\hat\bz_*^\intercal\hat\bSigma \hat\bz_*\right)^{-1/2}
\right),
\end{align}
with $\psi$ being the sigmoid function. Whereas GPC presents a computational cost of $\mathcal{O}(n^2)$ for each test instance, eq.~\eqref{eq:test} implies a complexity of $\mathcal{O}(D^2)$ in the case of our methods. In particular, notice that this is independent on the number of training instances. These significant reductions in computational cost (both at training and test) make our proposal suitable for large scale and real-time applications in general, and in EO applications in particular.

\begin{algorithm}
\caption{Training RFF-GPC and VFF-GPC.}
\label{algorithm}
\begin{algorithmic}
\REQUIRE Data set $\mathcal{D}=\{(\bx_i,y_i)\}_{i=1}^n \subset \R^d\times\{0,1\}$ and the number $D$ of Fourier frequencies.
\STATE Only for RFF-GPC, sample the Fourier frequencies $\bw_i$ from $\mathcal{N}(\mathbf{0},\bI)$ and fix them.

\STATE Initialize $\bxi^{(1)} = \mathbf{1}_{n\times 1}$, $\gamma^{(1)}=1$, and $\Phi^{(1)}$. For RFF-GPC (where $\Phi=\sigma$), $\Phi^{(1)}$ is initialized as the mean distance between (a subset of) the training data points $\bx_i$. For VFF-GPC (where $\Phi=(\sigma,\bW)$), $\sigma$ is initialized as described for RFF-GPC and $\bW$ with a random sample from $\mathcal{N}(\mathbf{0},\mathbf{I})$.
\REPEAT \STATE{Update $\bxi^{(k+1)}$ with eq.~\eqref{updateXi}.}
\STATE{Update $\Phi^{(k+1)}$ and $\gamma^{(k+1)}$ with eq.~\eqref{eq:Phi_gamma}, using $\Phi^{(k)}$ and $\gamma^{(k)}$ as initial values for the conjugate gradient method.}
\UNTIL{convergence}
\RETURN Optimal hyperparameter $\hat \Phi$ and the posterior distribution $\hat\p(\bbeta)=\mathcal{N}(\bbeta|\hat{\bmu},\hat{\bSigma})$.
\end{algorithmic}
\end{algorithm}

Finally, regarding the convergence of the proposed methods, we cannot theoretically guarantee the convergence to a global optimum (only a local one), since we are using conjugate gradient methods to solve the non-convex optimization problem in eq. \eqref{eq:Phi_gamma}.
However, from a practical viewpoint, we have experimentally checked that both methods have a satisfactory similar convergence pattern. Namely, in the first iterations, the hyperparameters experiment more pronounced changes, widely exploring the hyperparameters space. Then, once they reach a local optimum vicinity, these variations become smaller. Eventually, the hyperparameters values hardly change and the stop criterion is satisfied.

\section{Data Collection and Preprocessing}\label{sec:data}

This section introduces the datasets used for comparison purposes in the experiments. We considered (1) a continuous year of MSG data involving several hundred thousands of labeled pixels for cloud classification; and (2) a medium-size manually labeled dataset used to create the operational IASI cloud mask.

\subsection{Cloud detection over landmarks with Seviri/MSG}

We focus on the problem of cloud identification over landmarks using Seviri MSG data. This satellite mission constitutes a fundamental tool for weather forecasting, providing images of the full Earth disc every 15 minutes. Matching the landmarks accurately is of paramount importance in image navigation and registration models and geometric quality assessment in the Level 1 instrument processing chain. Detection of clouds over landmarks is an essential step in the MSG processing chain, as undetected clouds are one of the most significant sources of error in landmark matching (see Fig.~\ref{fig:motivation}). 

\begin{figure}[t!]
\centerline{\includegraphics[width=8.4cm]{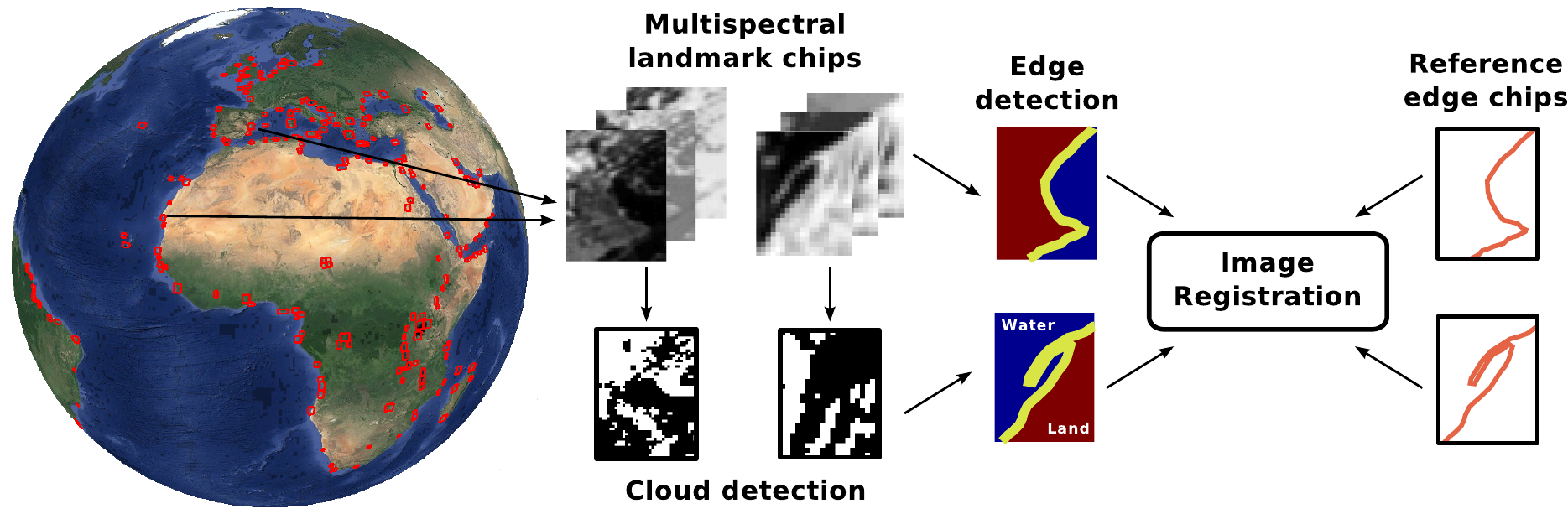}}
\vspace{-0.25cm}
\caption{Landmarks are essential in image registration and geometric quality assessment. Misclassification of cloud contamination in landmarks degrades the correlation matching, which is a cornerstone for the image navigation and registration algorithms.}
\label{fig:motivation}
\end{figure}

The dataset used in the experiments was provided by EUMETSAT, and contains Seviri/MSG Level 1.5 acquisitions for 200 landmarks of variable size for a whole year (2010). Landmarks mainly cover coastlines, islands, or inland waters. We selected all multispectral images from a particular landmark location, Dakhla (Western Sahara), which involves 35,040 MSG acquisitions with a fixed resolution of $20\times 26$ pixels. In addition, Level 2 cloud products were provided for each landmark observation, so the Level 2 cloud mask~\cite{Derrien05} is used as the best available `ground truth' to validate the results. We framed the problem for this particular landmark as a pixel-wise classification one.

A total amount of $16$ features were extracted from the images, involving band ratios, spatial, contextual and morphological features, and discriminative cloud detection scores. In particular, we considered: 7 channels converted to top of atmosphere (ToA) reflectance (R1, R2, R3, R4) and brightness temperature (BT7, BT9, BT10), 3 band ratios, and 6 spatial features. 
On the one hand, the three informative band ratios were: (i) a cloud detection ratio, ${\text{R}_{0.8\mu m}}/{\text{R}_{0.6\mu m}}$; (ii) a snow index, $({\text{R}_{0.6\mu m}-\text{R}_{1.7\mu m}})/({\text{R}_{0.6\mu m}+\text{R}_{1.7\mu m}})$; and (iii) the NDVI, $({\text{R}_{0.8\mu m}-\text{R}_{0.6\mu m}})/({\text{R}_{0.8\mu m}+\text{R}_{0.6\mu m}})$. On the other hand, the six spatial features were obtained by applying average filters of sizes $3\times 3$ and $5\times 5$, as well as a standard deviation filter of size $3\times 3$, on both bands R1 and BT9. 

Based on previous studies~\cite{Derrien05,Hocking10}, and in order to simplify the classification task, the different illumination conditions (and hence difficulty) over the landmarks are studied by splitting the day into four ranges (sub-problems) according to the solar zenith angle (SZA) values: \emph{high} (SZA$<$SZA$_{\rm median}$), \emph{mid} (SZA$_{\rm median}<$SZA$<80$\textdegree), \emph{low} (80\textdegree$<$SZA$<$90\textdegree), and \emph{night} (SZA$>$90\textdegree). Therefore, different classifiers are developed for each SZA range.

The final amount of pixels available for each illumination condition is $n=1500000$ for \emph{high}, \emph{mid}, and \emph{night}, and $n=1365083$ for \emph{low}. Moreover, each problem has different dimensionality: all the $d=16$ features were used for the three daylight problems, and $d=6$ was used for the night one (some bands and ratios are meaningless at night).

\subsection{Cloud detection with the IASI/AVHRR data}\label{sec:data_IAVISA}

The IAVISA dataset is part of the study ``IASI/AVHRR Visual Scenes Analysis and Cloud Detection'' (\url{http://www.brockmann-consult.de/iavisa-info-web/}), whose aim is to improve the IASI cloud detection by optimizing the coefficients used for a predefined set of cloud tests performed in the IASI Level 2 processing. The dataset was derived by visual analysis of globally distributed data, and served as input for the optimization and validation of the IASI cloud detection. Each collected IASI sample was classified concerning its cloudiness based on the visual inspection of the AVHRR Level 1B inside the IASI footprint. 
Each sample classifies a single IASI instantaneous field of view (IFOV) as being cloud-free (clear sky, 28\%), partly cloudy low (26\%), partly cloudy high (26\%), or cloudy (20\%). For the sake of simplicity, here we focus on discriminating between cloudy and cloud-free pixels.

\begin{figure}[t!]
\centerline{
\includegraphics[height=2.6cm]{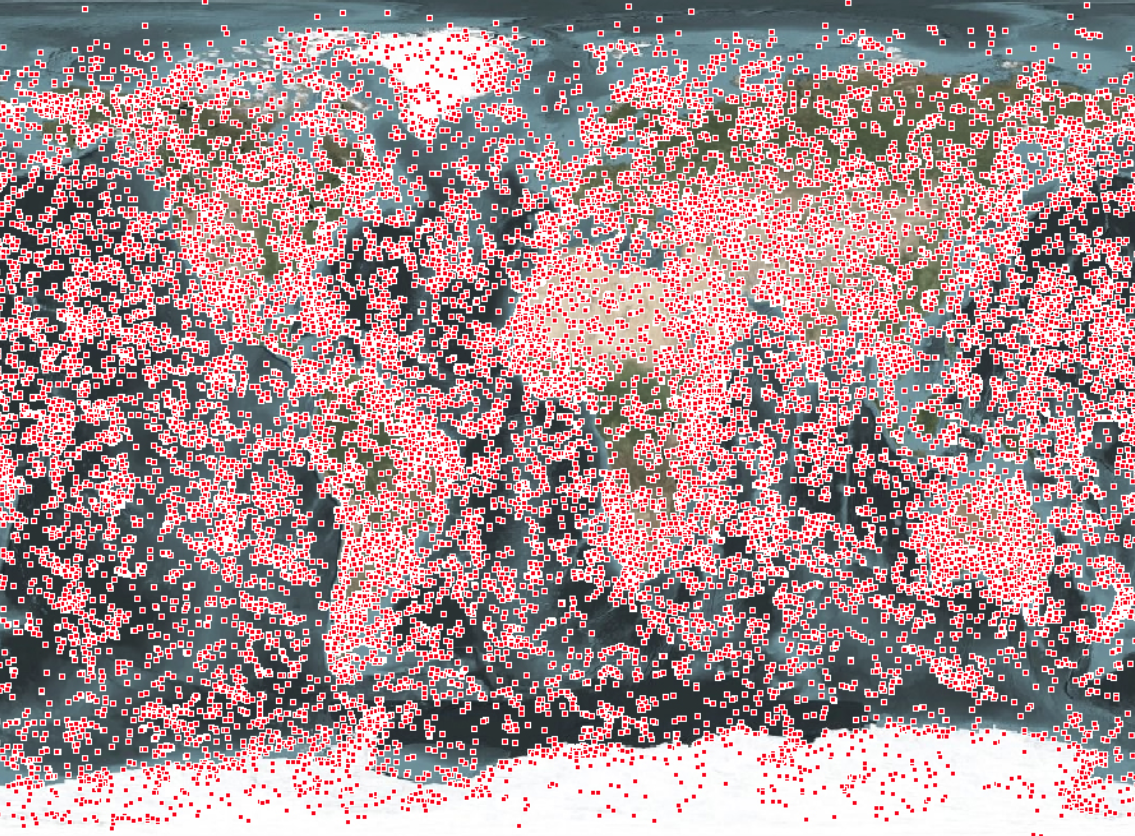}~
\includegraphics[height=2.6cm]{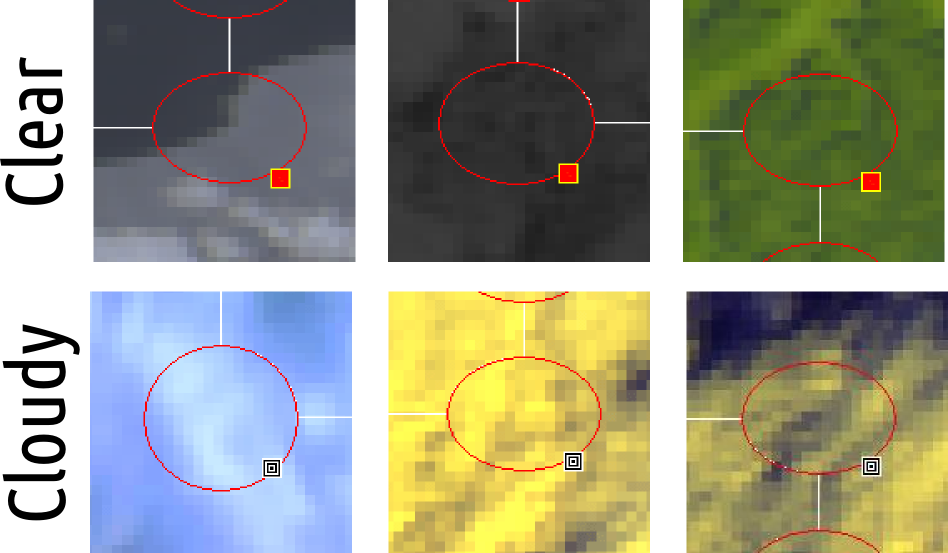}}
\vspace{-0.25cm}
\caption{Global sample coverage for all seasons, times of day, and cloud cases (left), and examples of cloud-free and cloudy samples (right).}
\label{fig:sampling}
\end{figure}

In order to ensure the representativeness of the dataset for the natural variability of clouds, labeling further considered additional conditions depending on: 1) the surface type (see Table~\ref{tab:IAVISA}), 2) the climate zone (K\"oppen classification over land, geographical bands over sea), 3) the season and 4) day/night discrimination. 
First, the surface type database used as ancillary information was the IGBP (International Geosphere-Biosphere Programme) scene types in the CERES/SARB (Clouds and the Earth's Radiant Energy System, Surface and Atmospheric Radiation Budget) surface map. The 18-class map was used to identify surface properties of a given region. The distribution of the surface types in the map is given in Table~\ref{tab:IAVISA}, showing the even distribution of clouds across land cover types which ensures representativeness (natural variability) of the database. Second, the different climate zones were sampled as follows: tropical ($n=7499$), dry ($n=3237$), temperate ($n=8150$), cold ($n=2205$), and polar zones ($n=3832$). Third, seasonality was also taken into account, and yielded the following distribution: Spring ($n=5862$), Summer ($n=6930$), Autumn ($n=6662$), and Winter ($n=5469$). 
The global sampling and some cloudy and cloud-free chips are shown in Fig.~\ref{fig:sampling}. The final database consists of $n=24923$ instances and $d=8461$ original features, which have been summarized to $d=100$ more informative directions through a standard Principal Component Analysis. 

\begin{table}[h!]
\small
\begin{center}
\caption{The samples distribution per surface types.} 
\label{tab:IAVISA} 
\begin{tabular}{|c|c|c|}
\hline
ID & Surface Type & Samples\\ \hline \hline
1 & Evergreen Needle Forrest & 603 \\ \hline
2 & Evergreen Broad Forrest & 876 \\ \hline
3 & Deciduous Needle Forrest & 89 \\ \hline
4 & Deciduous Broad Forrest & 324 \\ \hline
5 & Mixed Forest & 602 \\ \hline
6 & Closed Shrubs & 329 \\ \hline
7 & Open Shrubs & 1484 \\ \hline
8 & Woody Savannas & 768 \\ \hline
9 & Savannas & 656 \\ \hline
10 & Grassland & 886 \\ \hline
11 & Wetlands & 147 \\ \hline
12 & Crops &1294 \\ \hline
13 & Urban & 44 \\ \hline
14 & Crop/Mosaic & 1436 \\ \hline
15 & Snow/Ice & 1443 \\ \hline
16 & Barren/Desert & 1379 \\ \hline
17 & Water & 12150 \\ \hline
18 & Tundra & 413 \\ \hline
\end{tabular}
\end{center}
\end{table}

\section{Experiments}\label{sec:experiments}
In this section, we empirically demonstrate the performance of the proposed methods in the two real problems described above. Moreover, we carry out an exhaustive comparison to GPC with SE kernel (in the sequel, GPC for brevity).

In order to provide a deeper understanding of our methods, different values of $D$ (number of Fourier frequencies) will be used.
Different sizes $n$ of the training dataset will be also considered, in order to analyze the scalability of the methods and to explore the trade-off between accuracy and computational cost.
When increasing $n$ (respectively, $D$), we will add training instances (respectively, Fourier frequencies) to those already used for lower values of $n$ (respectively, $D$).
We provide numerical (in terms of recognition rate), computational (training and test times), and visual (by inspecting classification maps) assessments.  

\subsection{Cloud detection in the landmarks dataset}

\begin{figure*}[t!]
\begin{center}
\small
\setlength{\tabcolsep}{0.5pt}
\begin{tabular}{cccc}
\includegraphics[width=.25\textwidth]{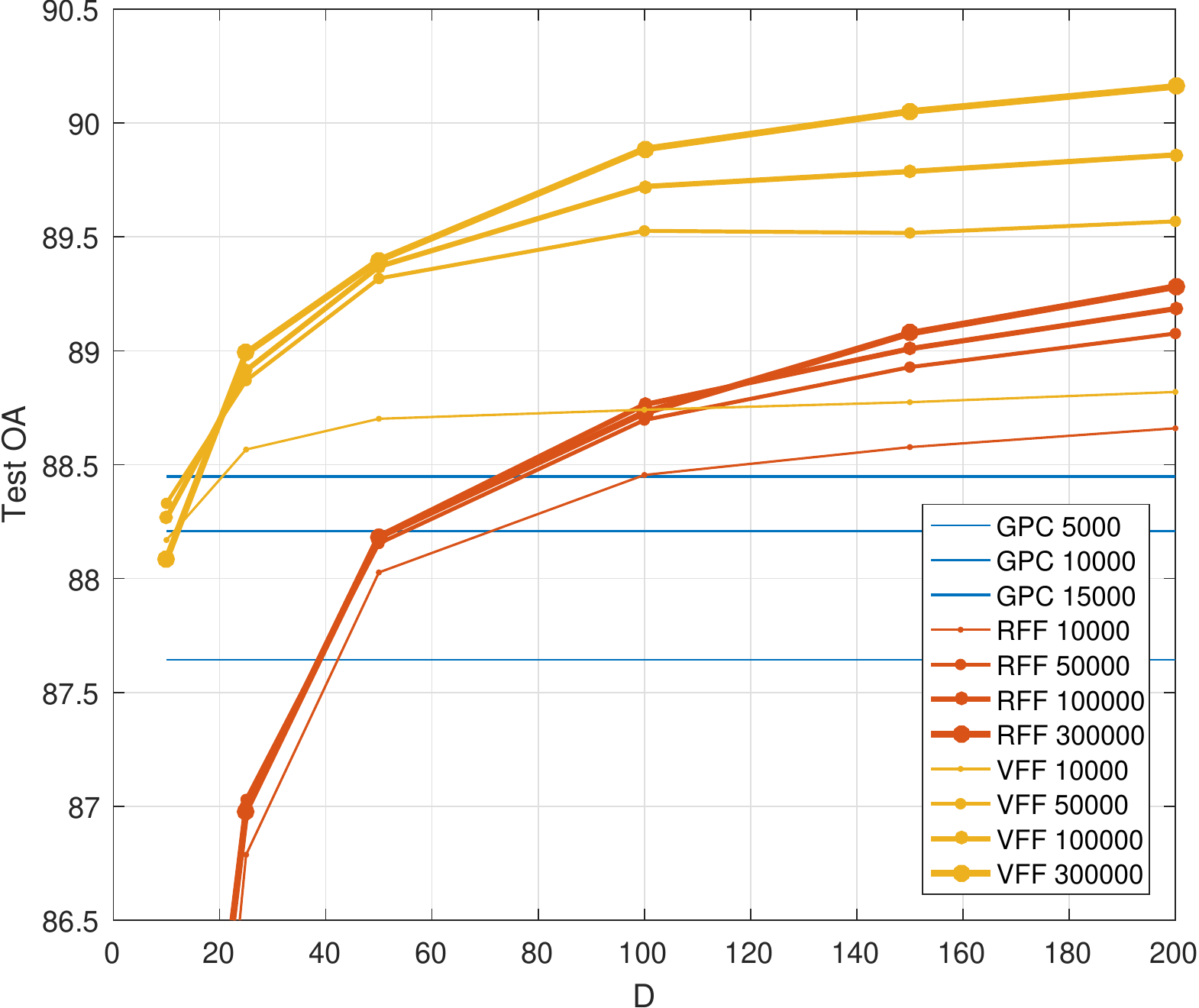}  &
\includegraphics[width=.25\textwidth]{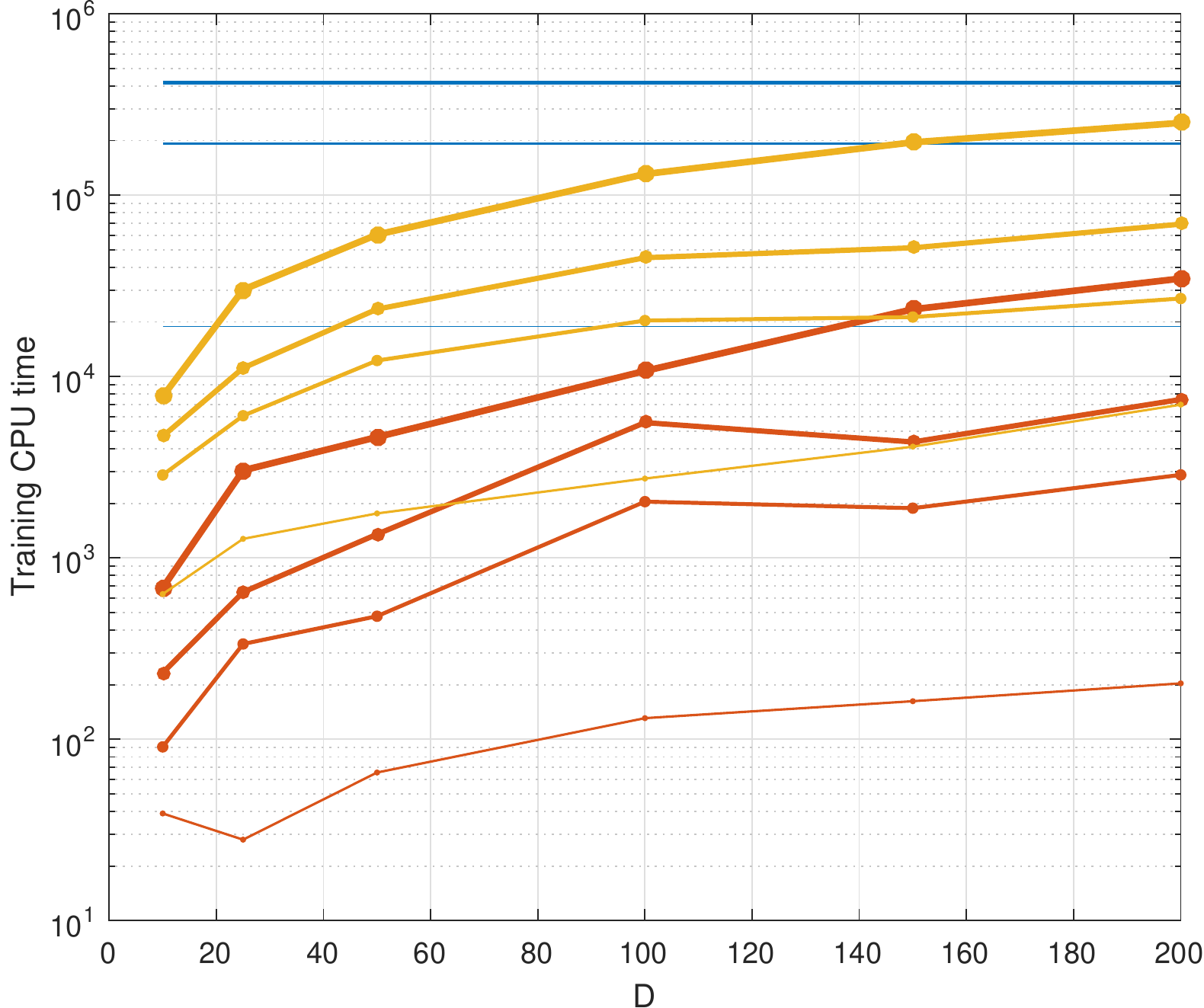}  &
\includegraphics[width=.25\textwidth]{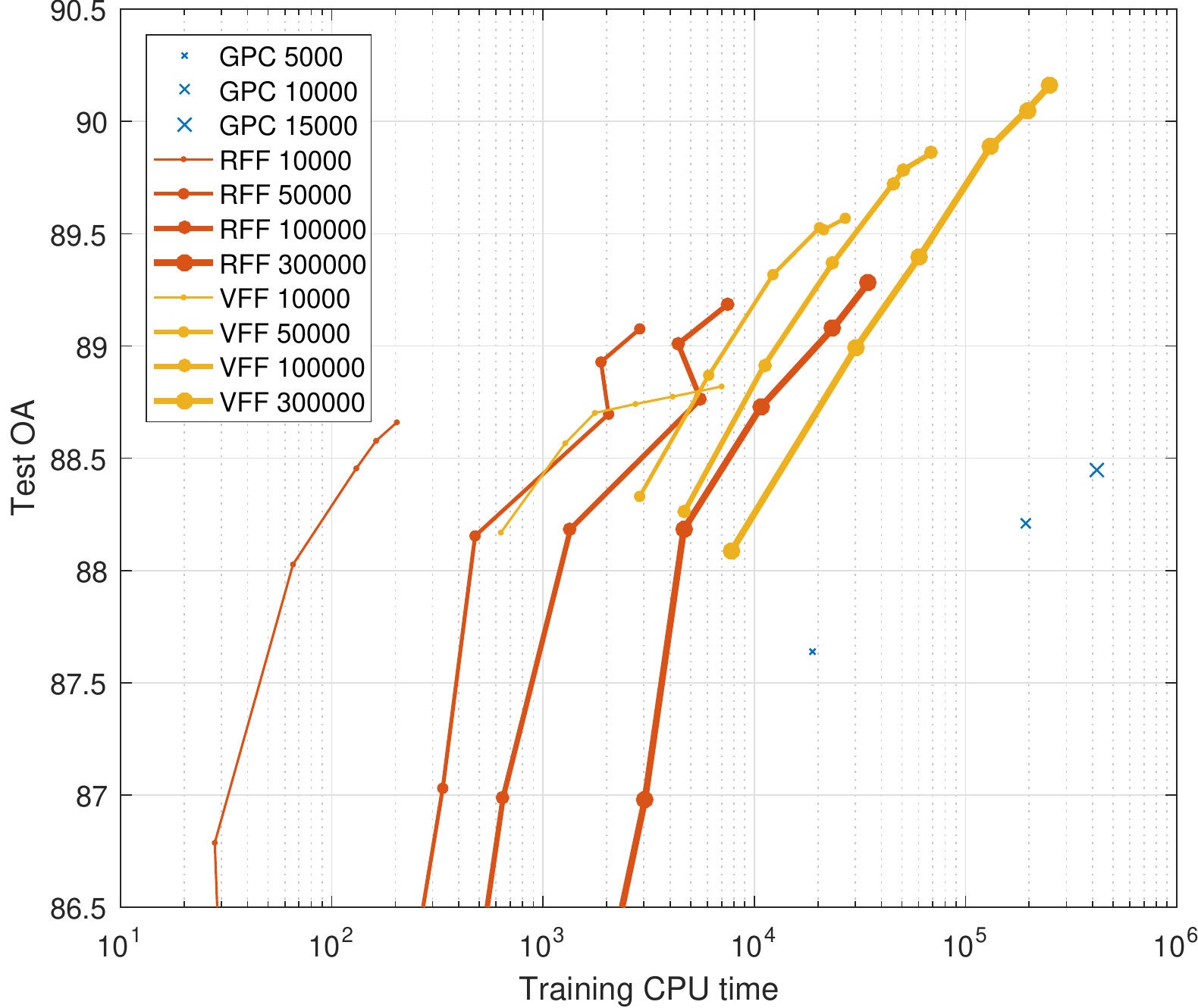} &
\includegraphics[width=.25\textwidth]{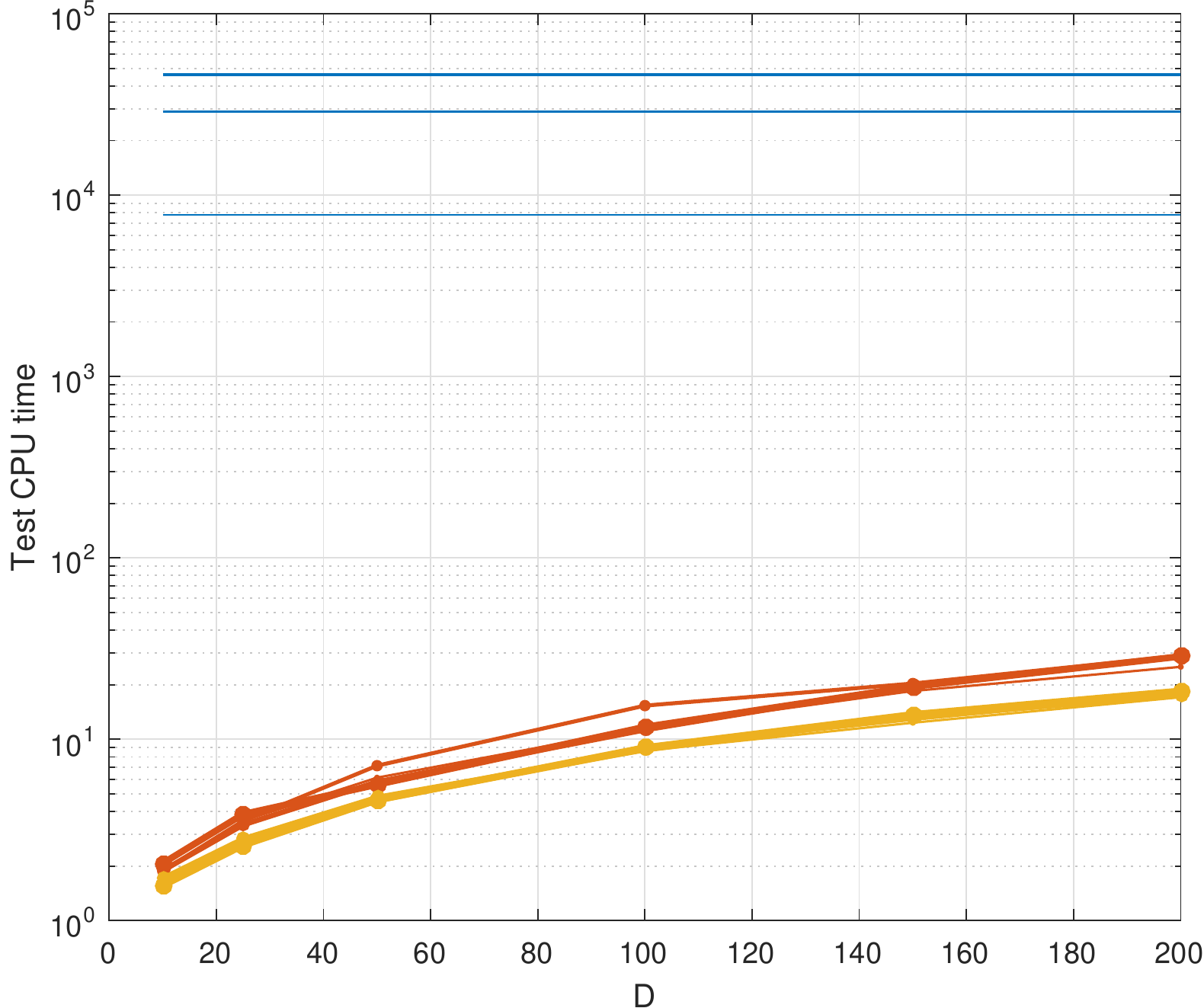}
\\
& & & \\
\includegraphics[width=.25\textwidth]{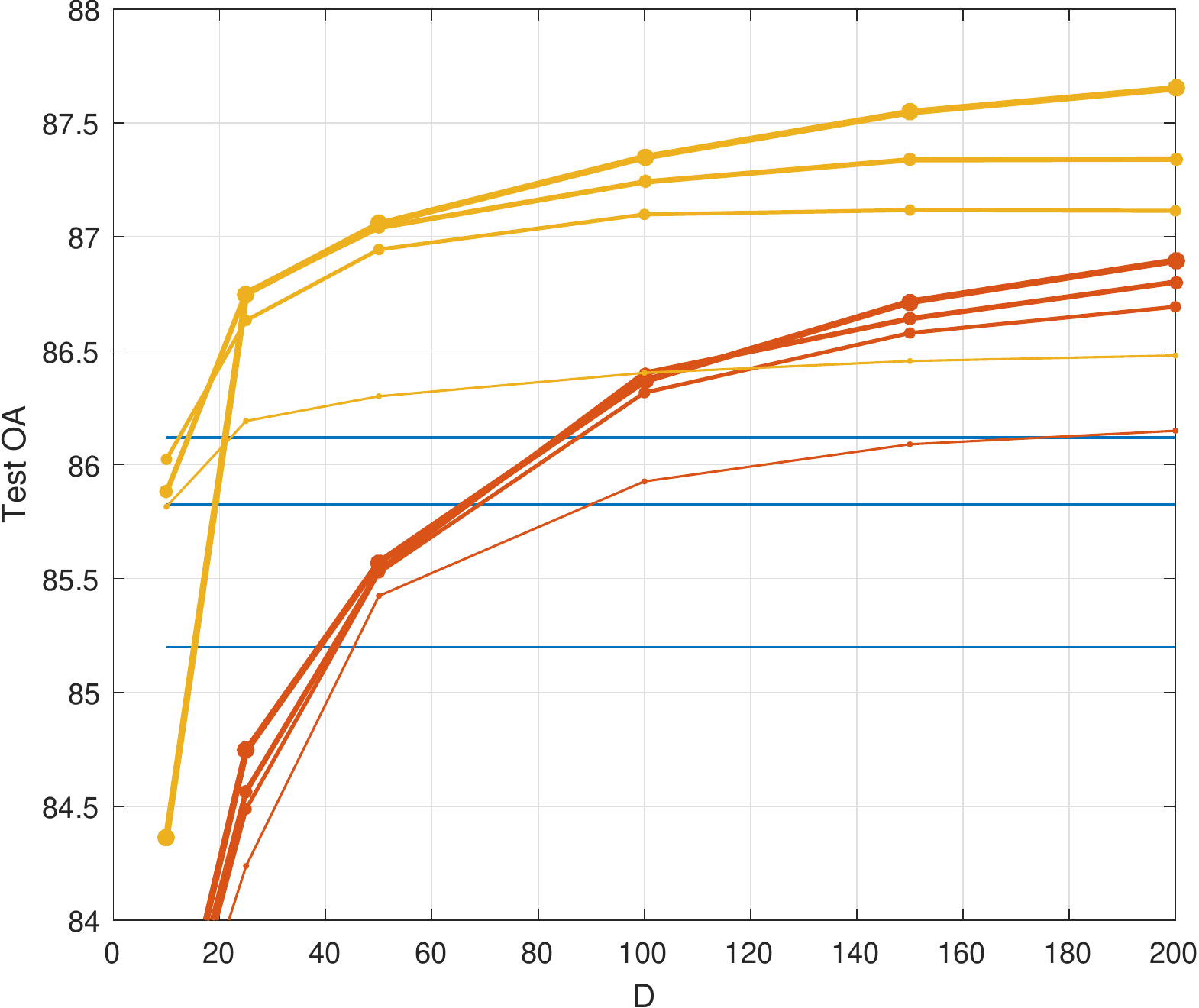}  &
\includegraphics[width=.25\textwidth]{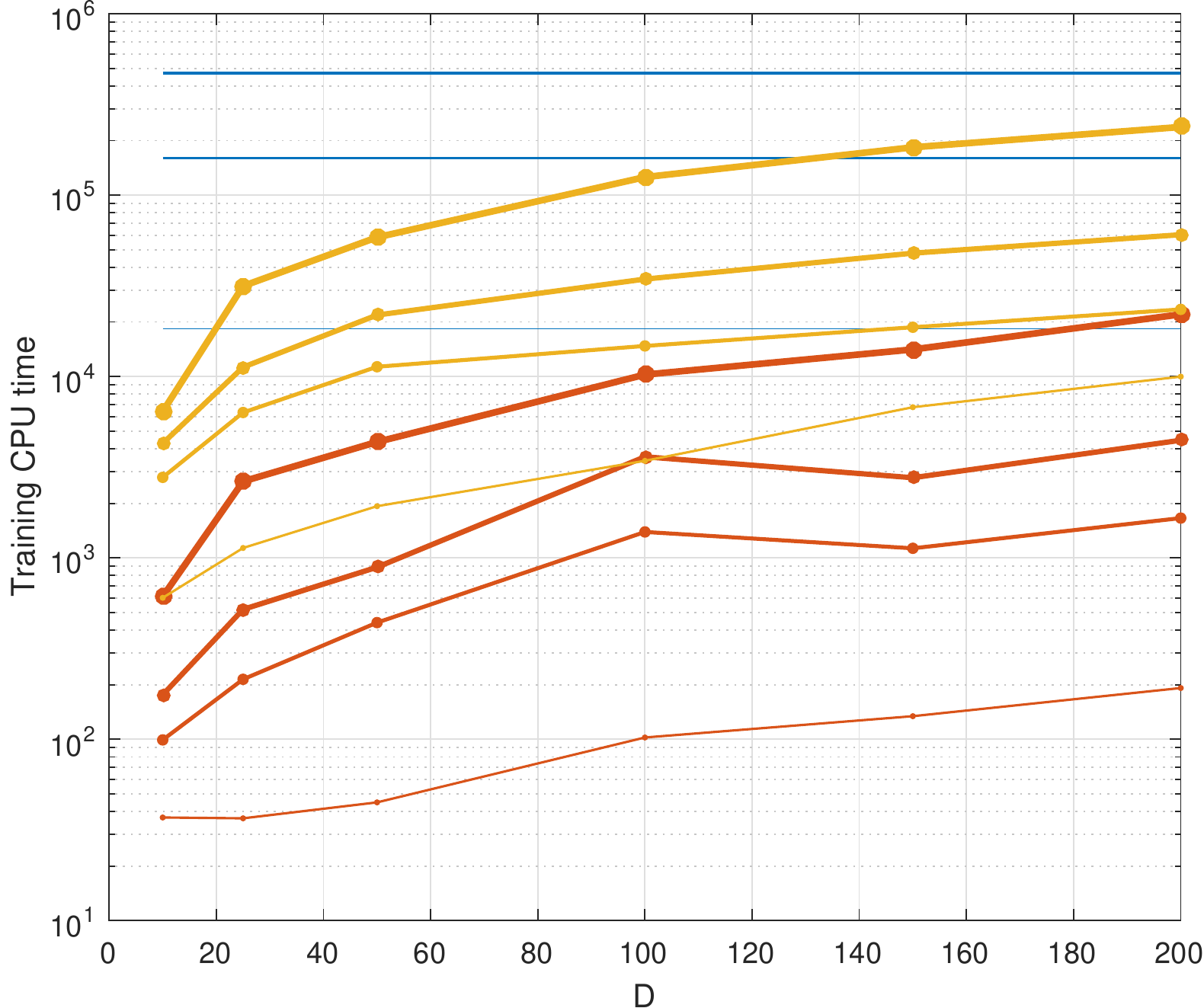}  &
\includegraphics[width=.25\textwidth]{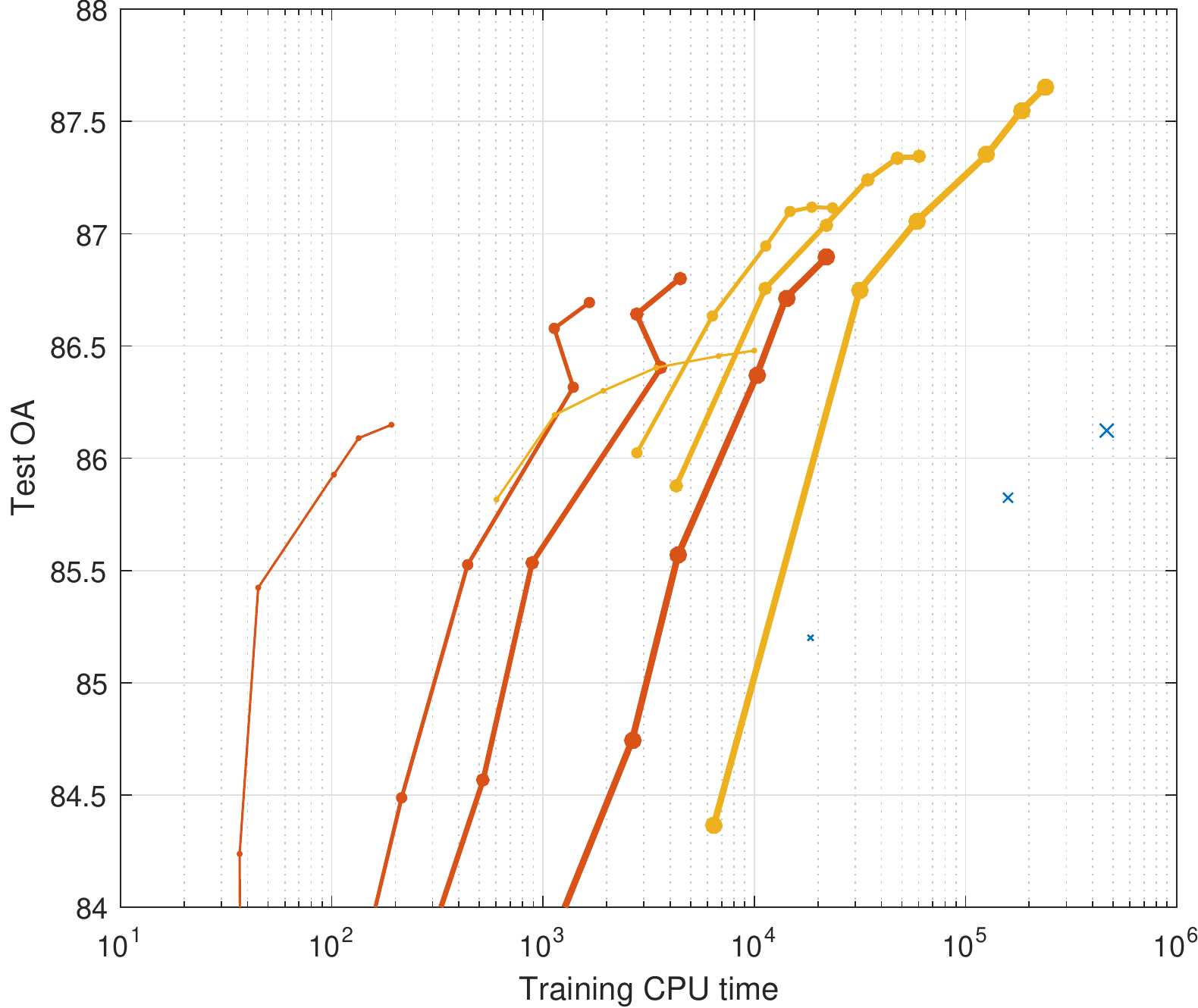} &
\includegraphics[width=.25\textwidth]{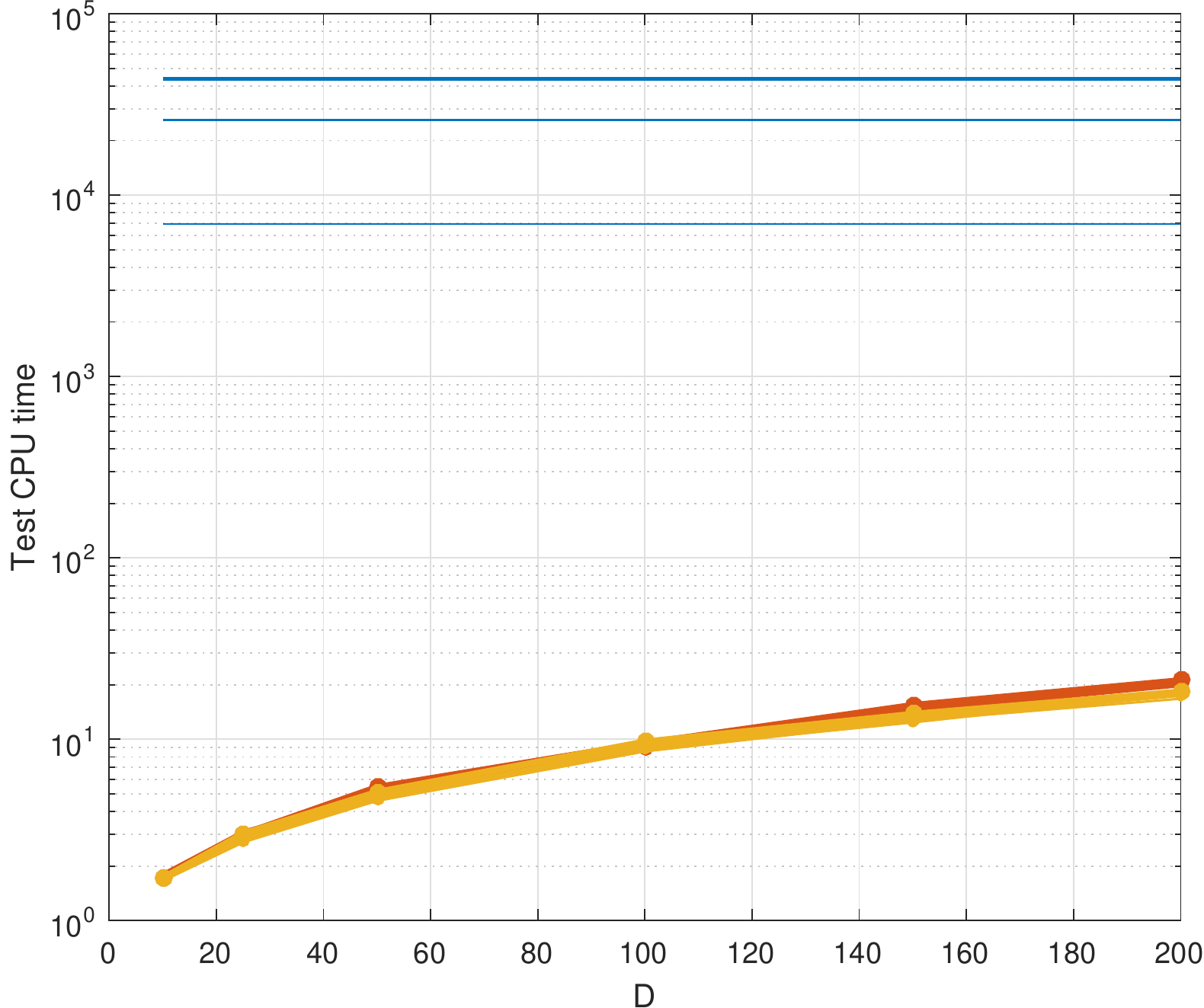}
\\
& & & \\
\includegraphics[width=.25\textwidth]{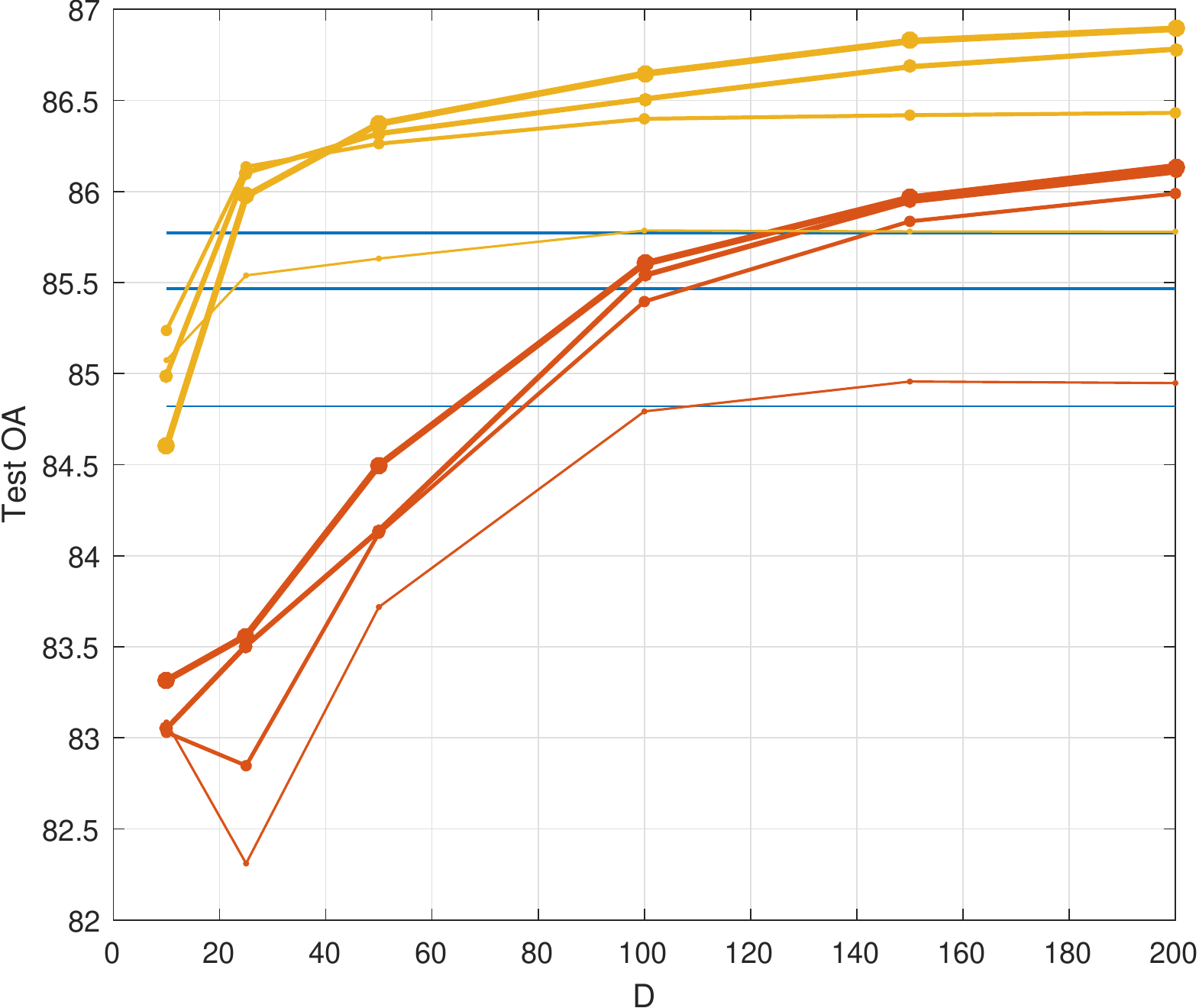}  &
\includegraphics[width=.25\textwidth]{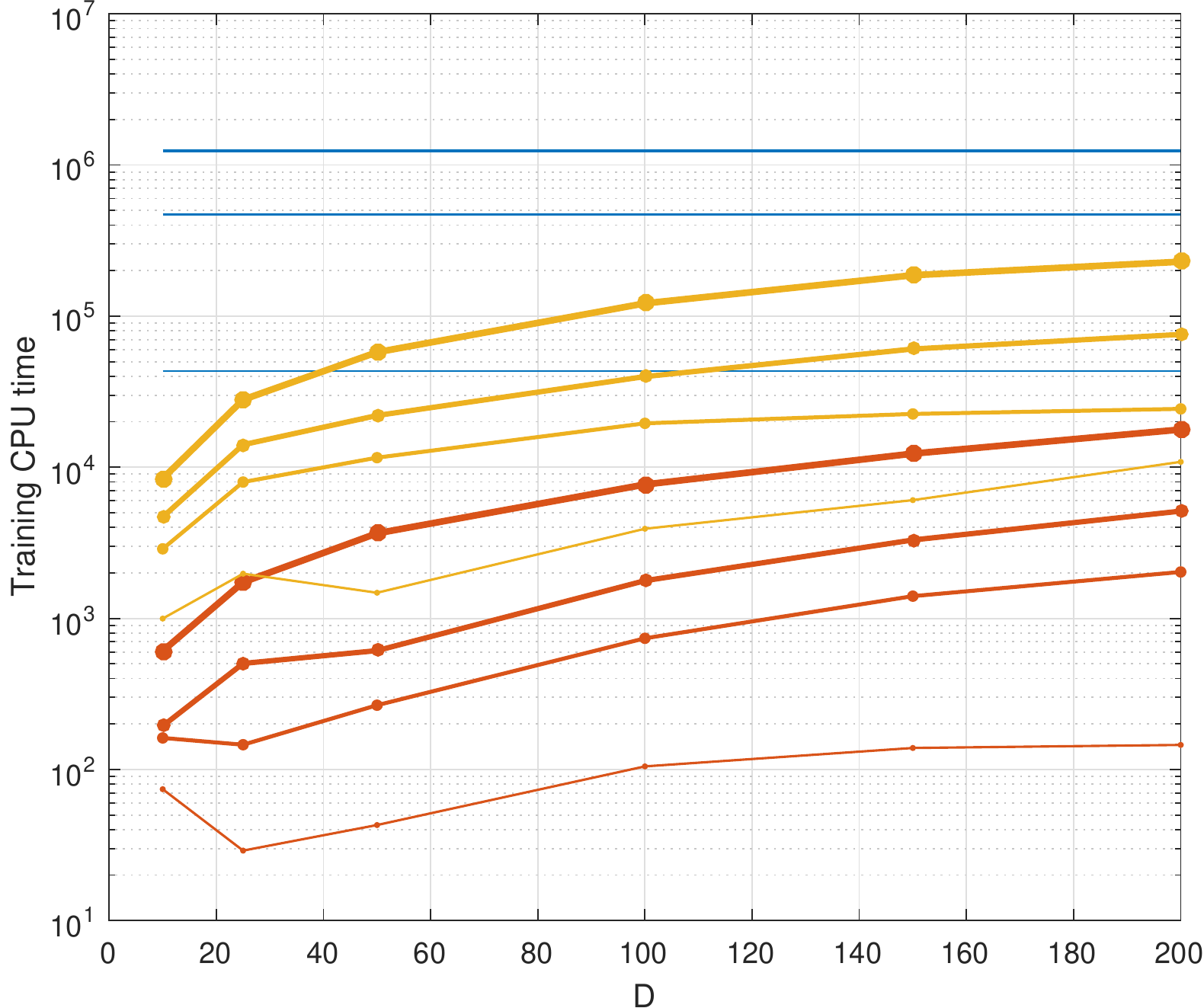}  &
\includegraphics[width=.25\textwidth]{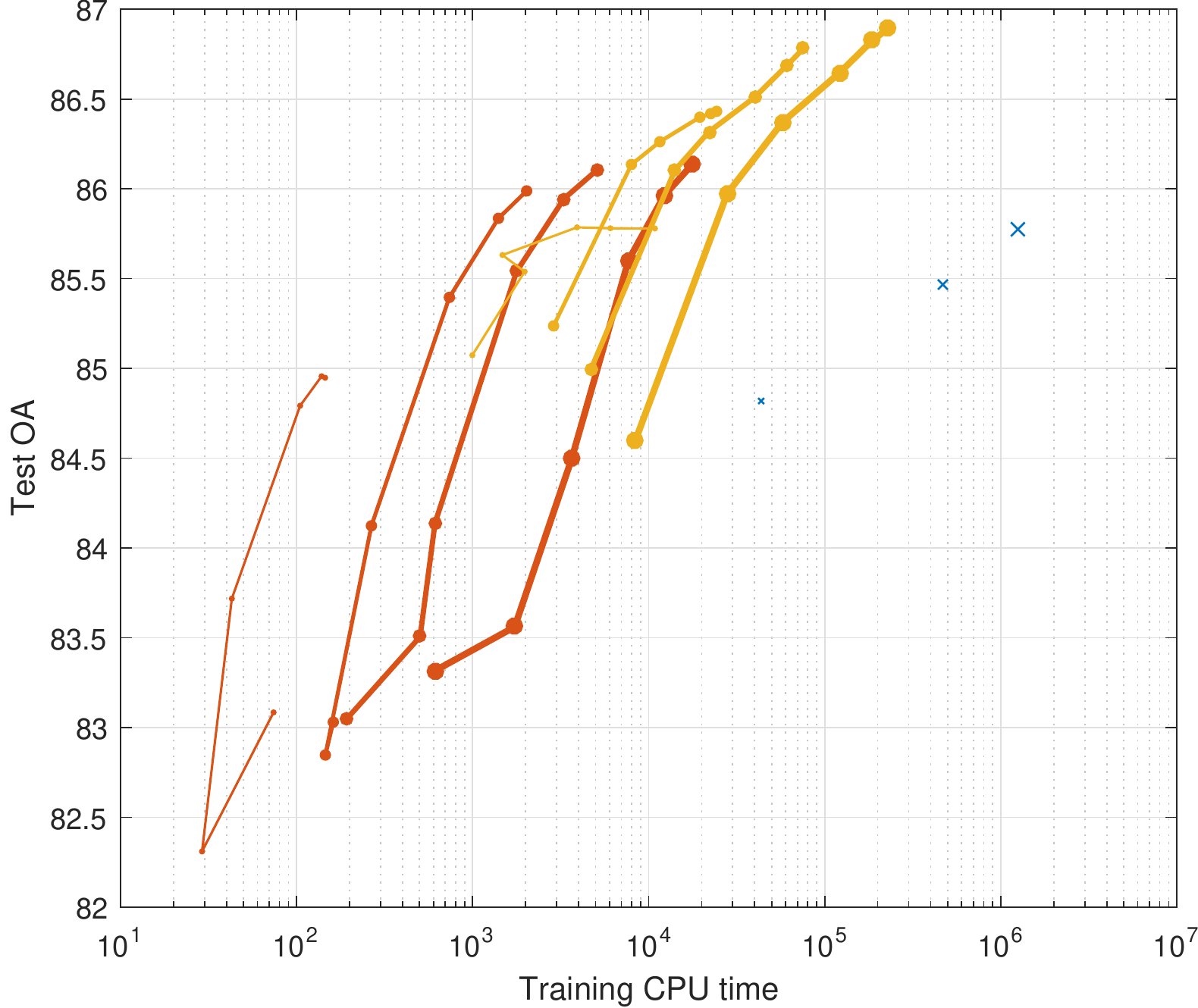} &
\includegraphics[width=.25\textwidth]{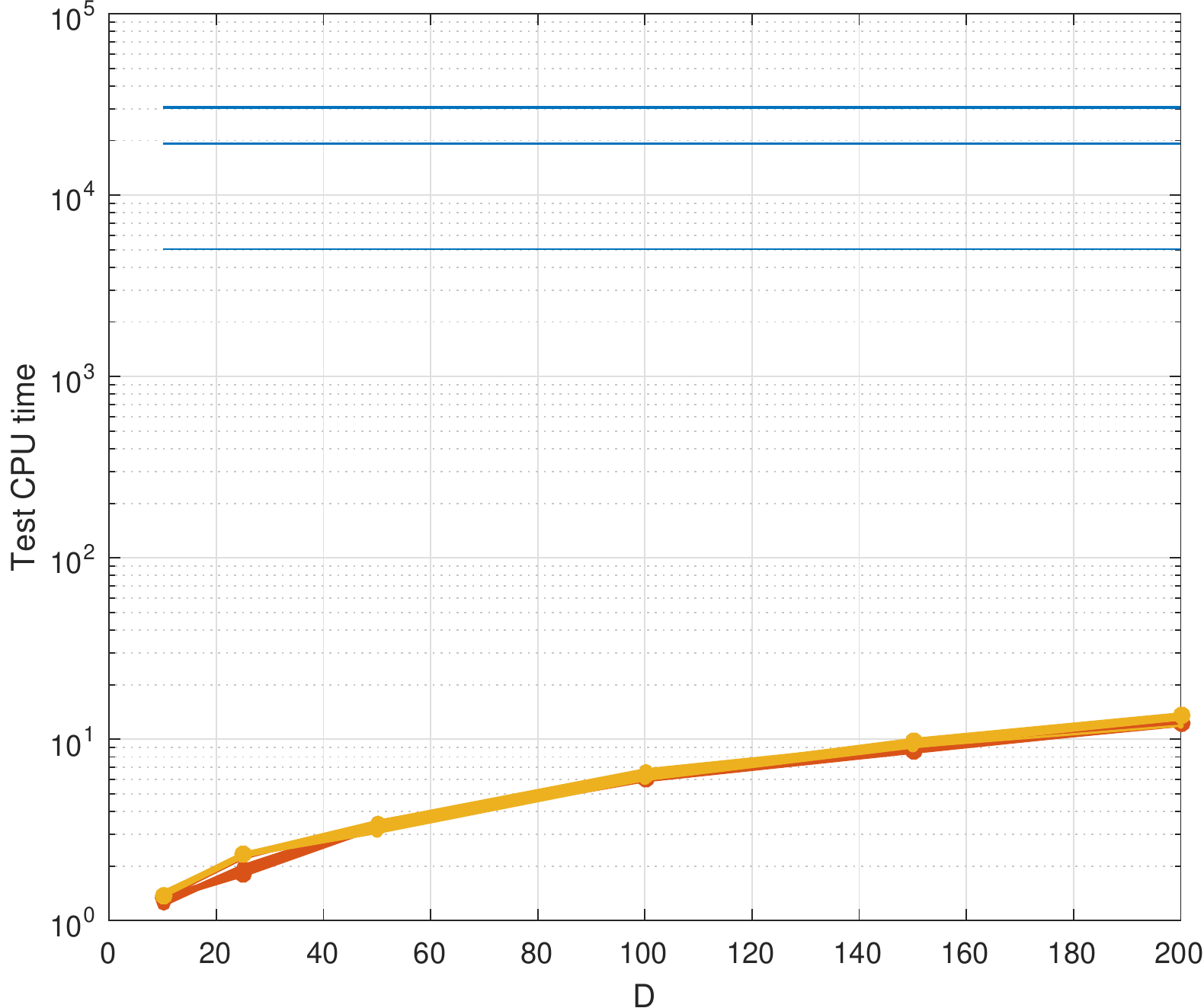}
\\
& & & \\
\includegraphics[width=.25\textwidth]{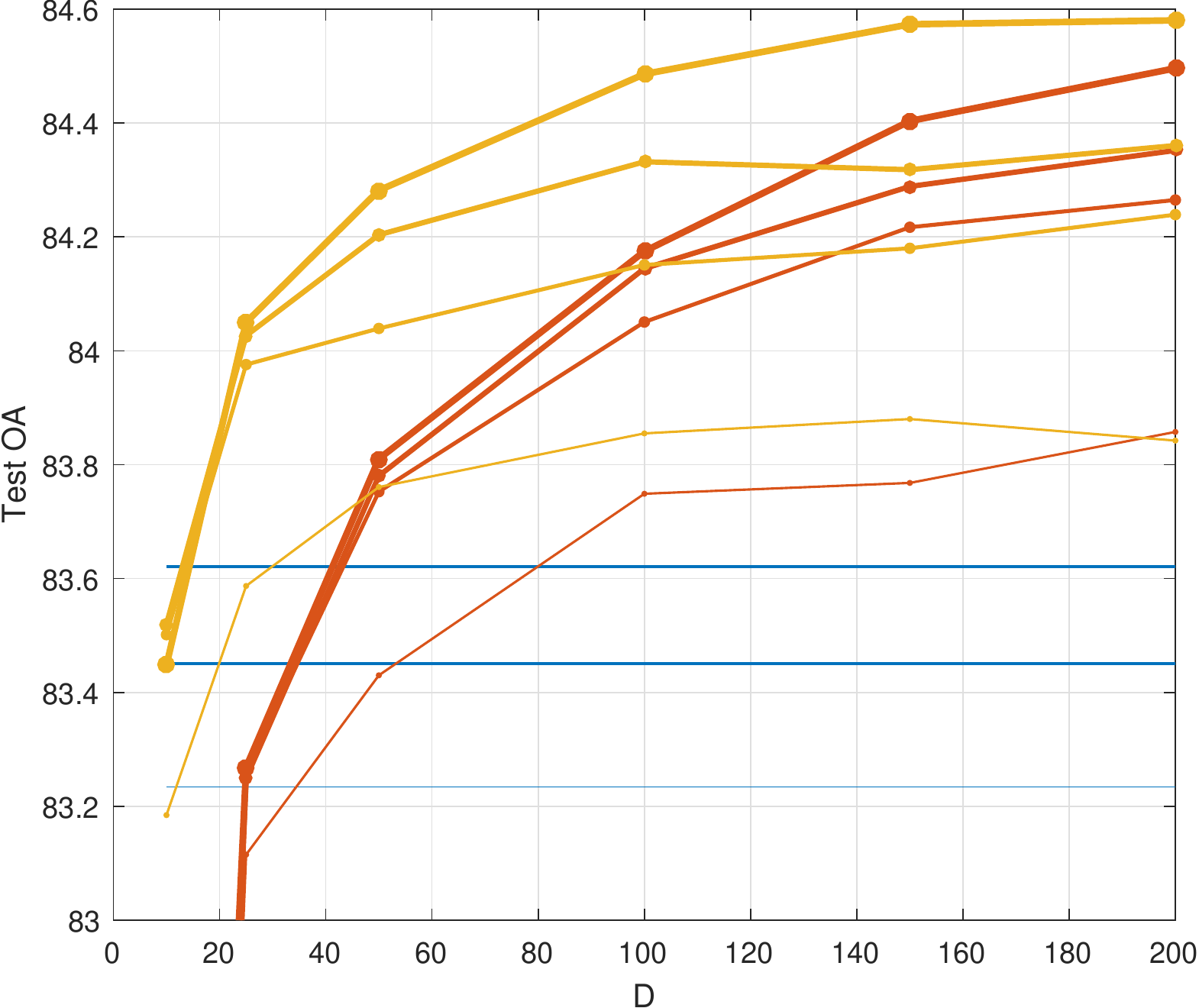}  &
\includegraphics[width=.25\textwidth]{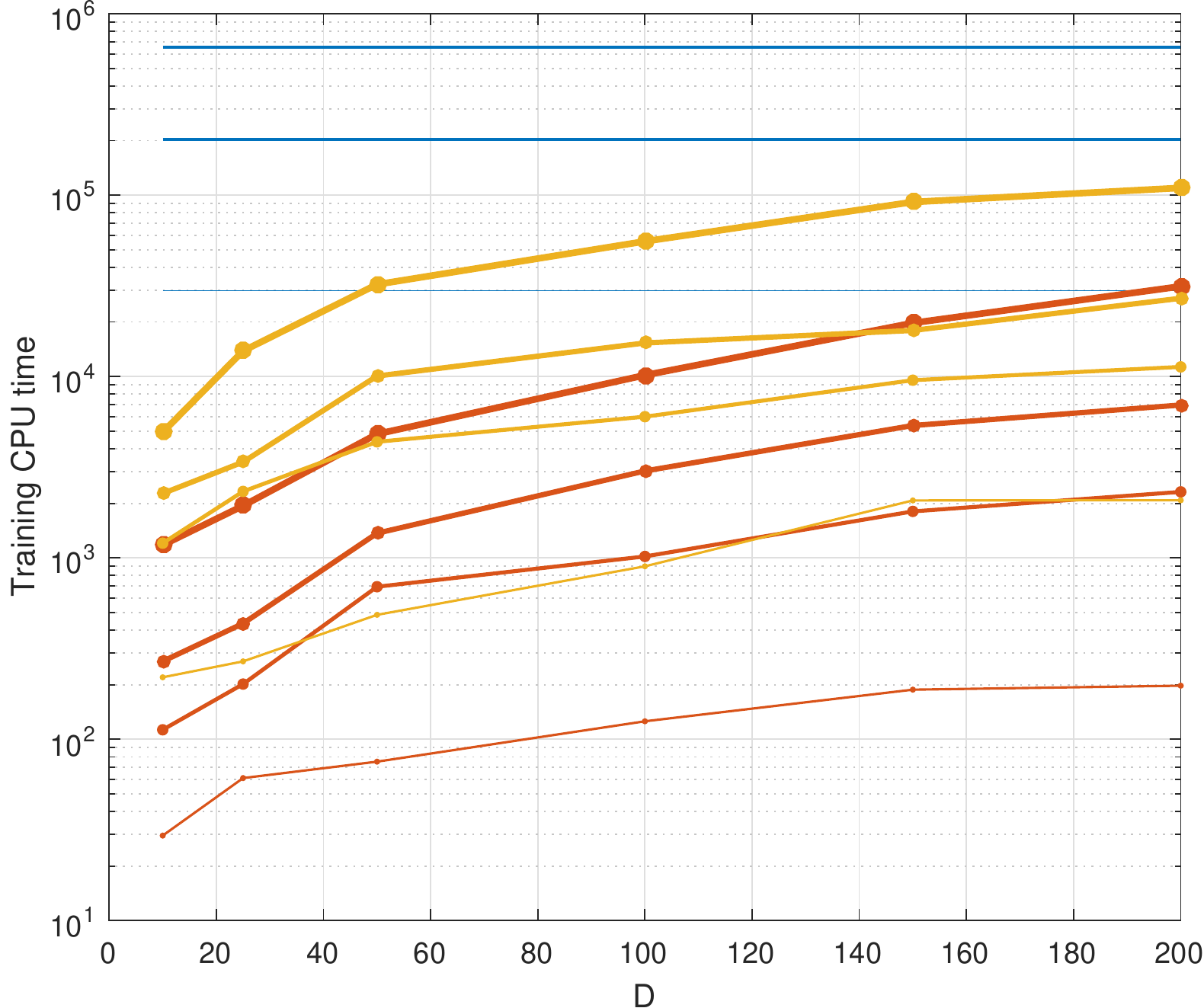}  &
\includegraphics[width=.25\textwidth]{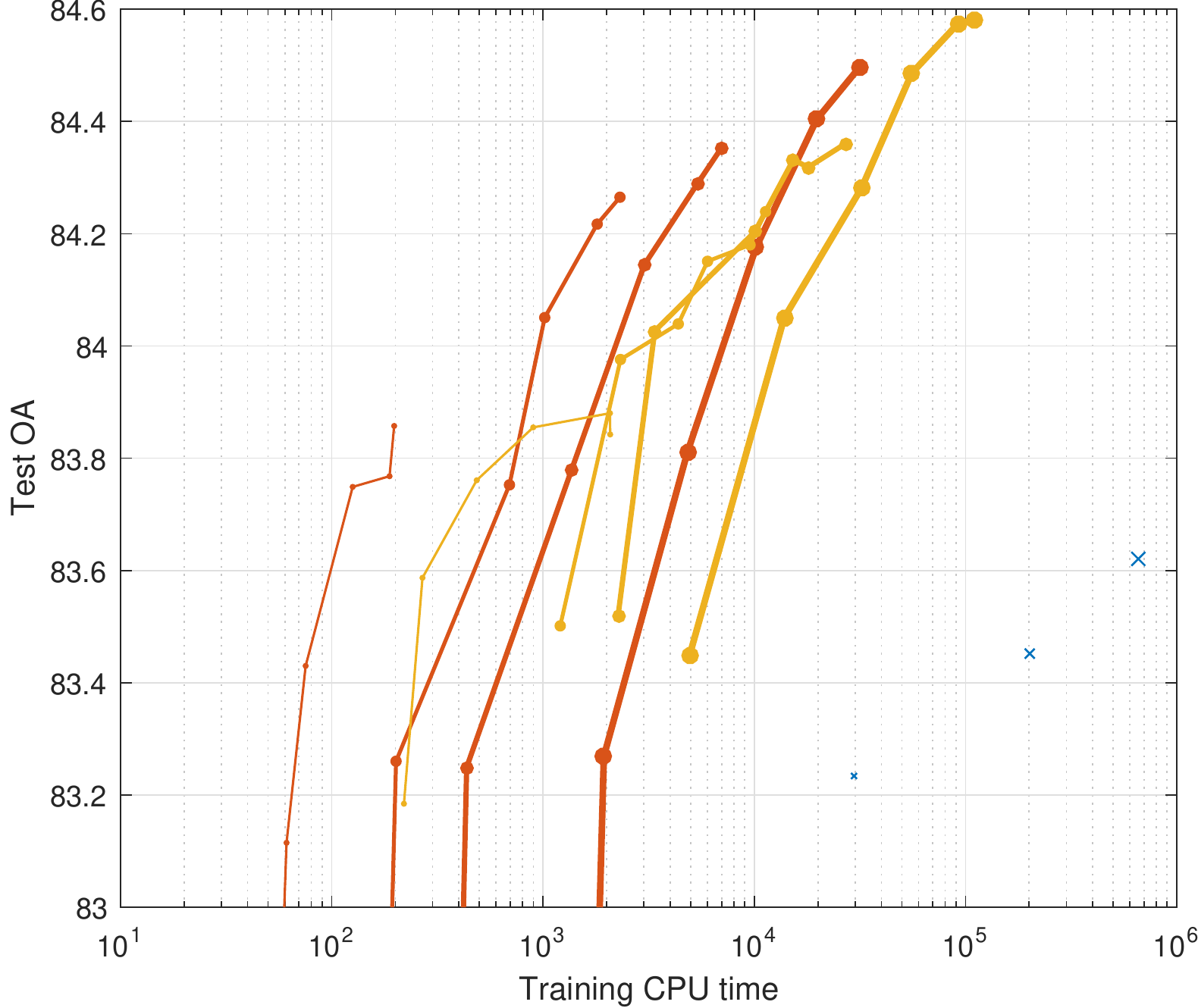}  &
\includegraphics[width=.25\textwidth]{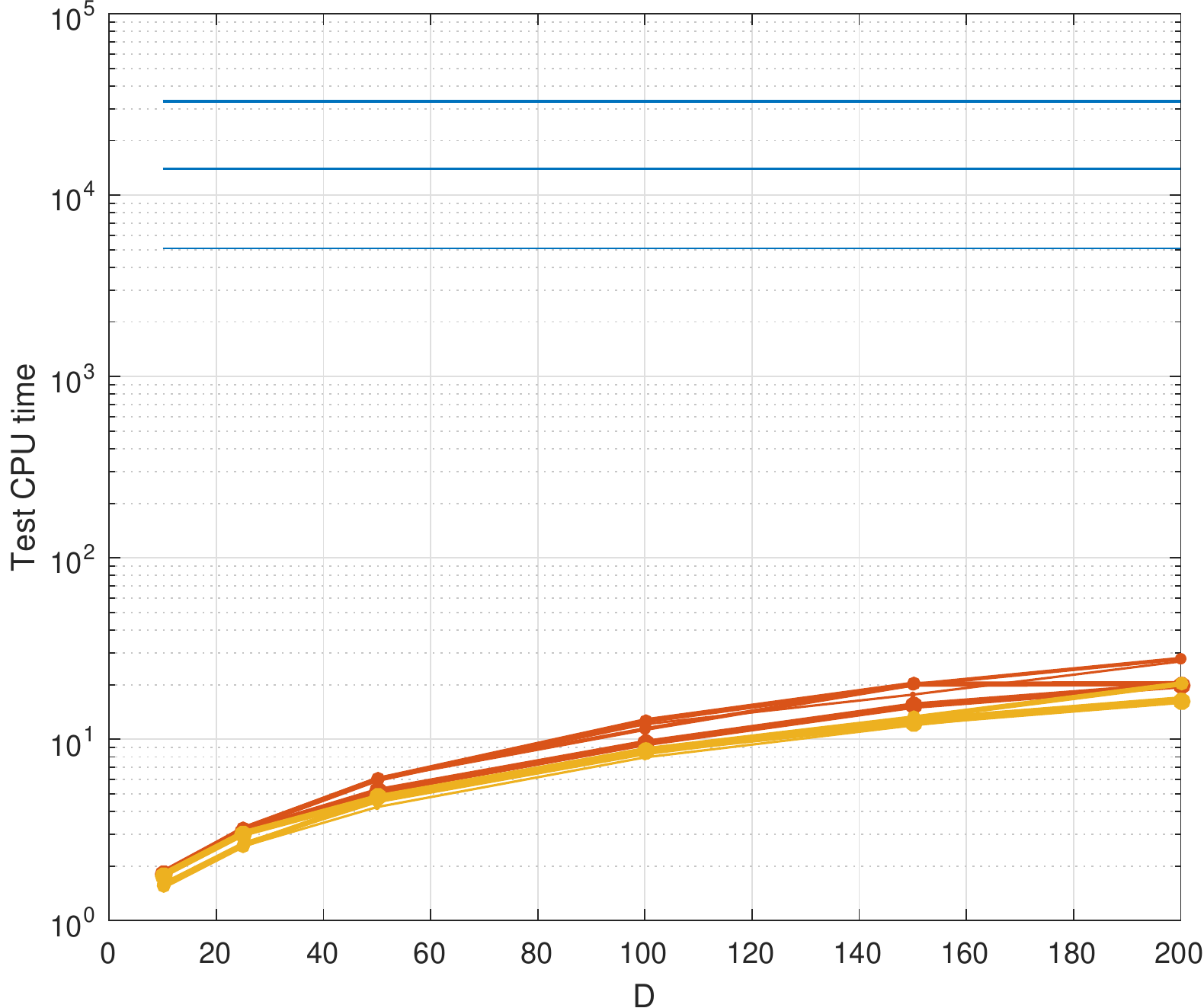}
\end{tabular}
\vspace{-0.25cm}
\caption{Experimental results for LANDMARKS dataset. From top to bottom, the rows correspond with the described \emph{high}, \emph{mid}, \emph{low}, and \emph{night} illumination conditions. For each row, the first column shows the test overall accuracy (OA) of RFF-GPC, VFF-GPC, and GPC for the different values of $n$ (number of training examples) and $D$ (number of Fourier frequencies) considered. 
The second column is analogous, but displays the CPU time (in seconds) needed to train each method (instead of the test OA).
The third column summarizes the two previous ones, providing a trade-off between test OA and training CPU time.
The last column is analogous to the first and second ones, but showing the CPU time used at the test step (production time).
The legend for the second and fourth columns is the same as in the first one. However, notice that in the third column plots the GPC lines degenerate into single points (since GPC does not depend on $D$). In both legends, the numbers indicate the amount $n$ of training examples used, which determines the width/size of the lines/points too.
As further explained in the main text, shown results are the mean over five independent runs.}
\label{fig:LANDMARKS_main}
\end{center}
\end{figure*}

In this large scale problem, the number of training examples is selected as $n\in\{10000,50000,100000,300000\}$ for RFF-GPC and VFF-GPC, and $n\in\{5000,10000,15000\}$ for GPC.
Notice that the improvement at training computational cost ($\mathcal{O}(nD^2)$ for the proposed methods against $\mathcal{O}(n^3)$ for GPC) enables us to consider much greater training datasets for our methods.
In fact, as we will see in Figure \ref{fig:LANDMARKS_main}, even with $n=300000$ RFF-GPC and VFF-GPC are computationally cheaper than GPC with $n=15000$.
Indeed, higher values of $n$ are not considered for GPC to avoid exceeding the (already expensive) $10^6$ seconds of training CPU time needed with just $n=15000$. Regarding the number $D$ of Fourier frequencies, we use $D\in\{10,25,50,100,150,200\}$.

The experimental results, which include predictive performance (test overall accuracy), training CPU time, and test CPU time, are shown in Figure \ref{fig:LANDMARKS_main}. 
Every single value is the average of five independent runs under the same setting.
Namely, for each illumination condition, a test dataset is fixed and five different balanced training datasets are defined with the remaining data.
Notice that a general first observation across Figure \ref{fig:LANDMARKS_main} 
suggests that higher accuracies are obtained for higher illumination conditions (SZA).



Figure \ref{fig:LANDMARKS_main} reveals an overwhelming superiority of RFF-GPC and VFF-GPC over standard GPC: our proposed methods achieve a higher predictive performance while investing substantially lower training CPU time. This is very clear from the third column plots, where for any blue point we can find orange and yellow points which are placed more north-west (i.e. higher test OA and lower CPU training time). Furthermore, the fourth column shows an equally extraordinary reduction in test CPU time (production time), where the proposed methods are more than $100$ times faster than GPC. In particular, this makes RFF-GPC and VFF-GPC better suited than standard GPC for real-time EO applications.

Regarding the practical differences between RFF-GPC and VFF-GPC, we observe that RFF-GPC is faster (at training) whereas VFF-GPC is more accurate. This is a natural consequence of their theoretical formulations: the estimation of the Fourier frequencies $\bW$ in VFF-GPC makes it more flexible and expressive, but involves a heavier training.
Therefore, in this particular problem, the final practical choice between the two proposed methods would depend on the relative importance that the user assigns to test accuracy (where VFF-GPC stands out) and training cost (where RFF-GPC does so). 
In terms of test cost, both methods are very similar, as expected from the identical $\mathcal{O}(D^2)$ theoretical test complexity. The independence of this quantity on $n$ is also intuitively reflected in the experiments, with all the RFF-GPC and VFF-GPC lines collapsing onto a single one in the fourth column plots of Figure \ref{fig:LANDMARKS_main}.

At this point, it is worth to analyze a bit further the role of $D$ in the performance of our methods. Recall (Section \ref{sec:theory_RFF}) that RFF-GPC is an approximation to GPC, with an error that exponentially decreases with the ratio $D/d$ between the dimensions of the projected Fourier features space and the original one. Therefore, it is theoretically expected that the performance of RFF-GPC increases with $D$, becoming equivalent to GPC when $D\rightarrow\infty$. Actually, this is supported by the first column of Figure \ref{fig:LANDMARKS_main}. Moreover, since our problem here presents a low $d$ ($16$ for \emph{high}, \emph{mid}, and \emph{low}, and $6$ for \emph{night}), it is natural that RFF-GPC with just $D=200$ already gets very similar (even better in some cases) results to standard GPC with the same $n$ (yet far much faster, compare GPC and RFF/VFF for $n=10000$).
In the case of VFF-GPC, where the Fourier frequencies are model parameters to be estimated, the number $D$ is directly related to the complexity of the model. Therefore, its increase should not always mean a higher performance in test OA, since large values may provoke over-fitting to the training dataset (this will be clear in the next dataset, whereas it does not occur in LANDMARKS). Furthermore, unlike RFF-GPC, the performance of VFF-GPC is not directly affected by $d$.

It is also reasonable to expect that both test OA and training CPU time increase with the training dataset size $n$. More specifically, and from a practical perspective in which the computational resources are finite, the first column in Figure \ref{fig:LANDMARKS_main} shows that test OA becomes stalled when only one of $n$ or $D$ increases. However, greater improvements in test OA are achieved when $n$ and $D$ are jointly increased. Notice that this is also justifiable from a theoretical viewpoint: the higher the dimensionality of the projected Fourier features space (which is $2D$), the larger number $n$ of examples are required to identify the separation between classes.


\if false
\subsection{Computational cost}
\begin{figure}[h!]
\centerline{\includegraphics[width=8.4cm]{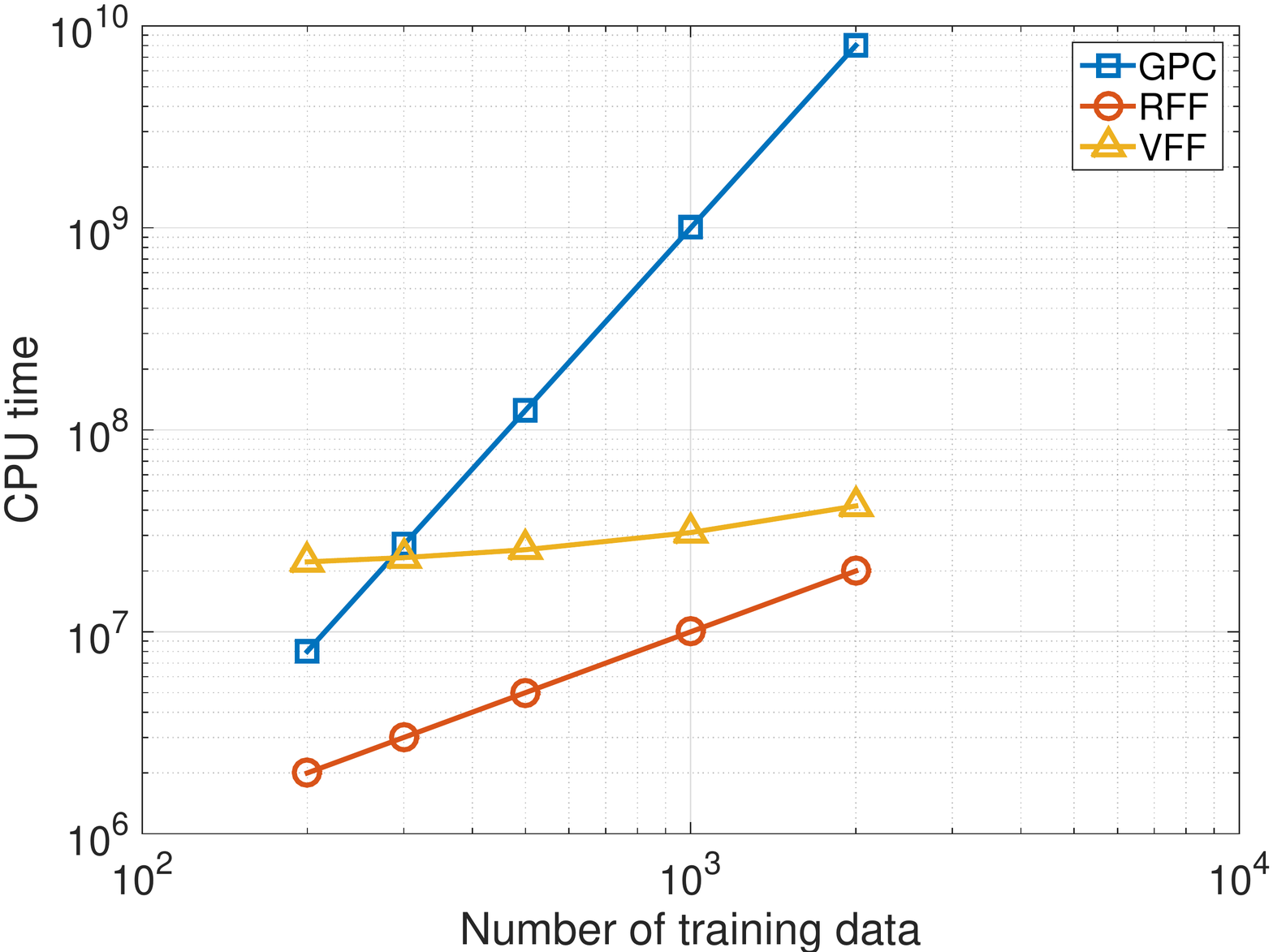}}
\vspace{-0.25cm}
\caption{Landmarks are essential in image registration and geometric quality assessment. Any misclassification of a landmark due to cloud contamination degrades the correlation matching which is a cornerstone for the image navigation and registration (INR) algorithms.}
\label{fig:motivation}
\end{figure}
\fi



\subsection{Cloud detection with the IAVISA dataset}

\begin{figure*}[t!]
\begin{center}
\small
\setlength{\tabcolsep}{7pt}
\begin{tabular}{cc}
\includegraphics[width=.35\textwidth]{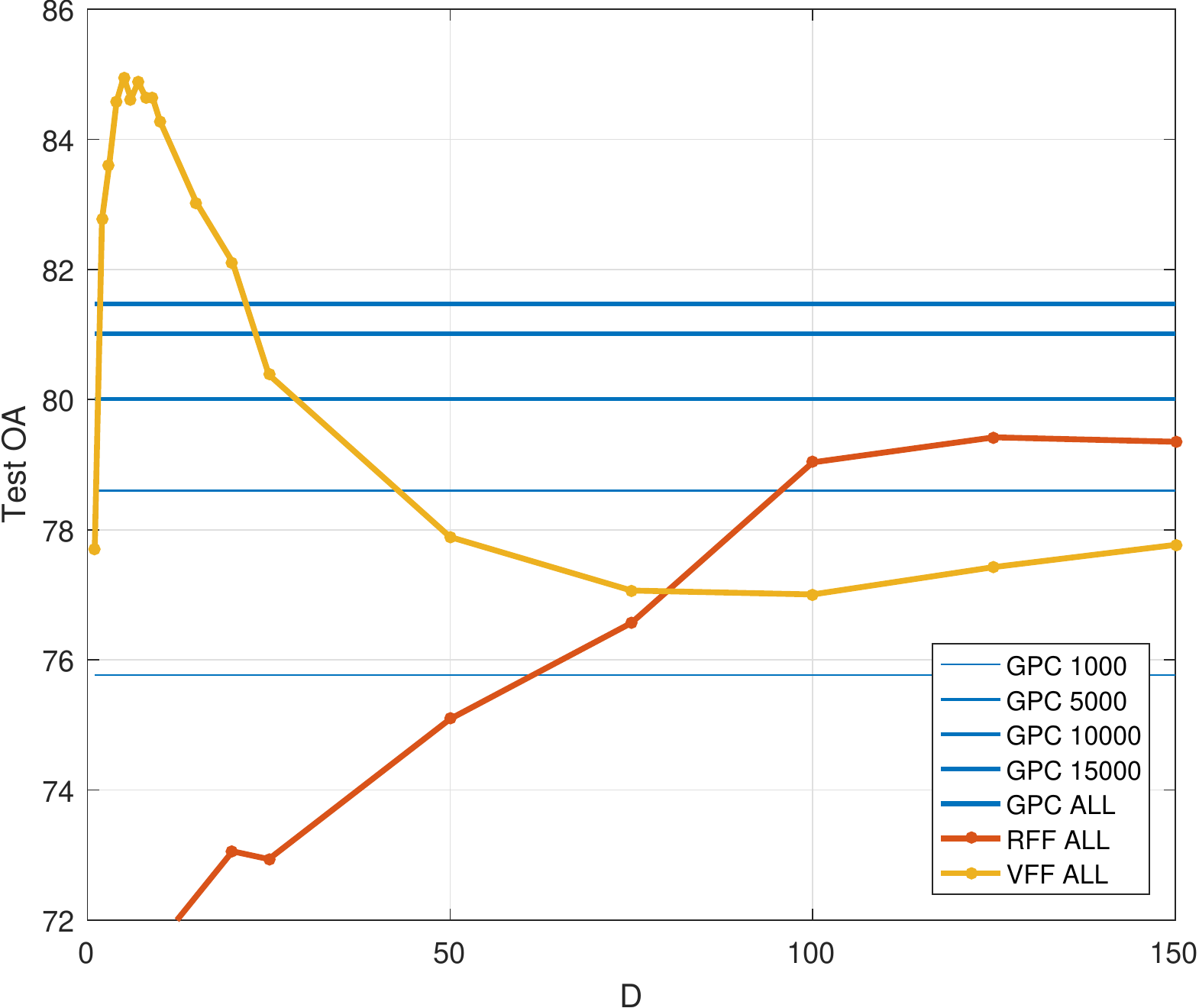}
&
\includegraphics[width=.35\textwidth]{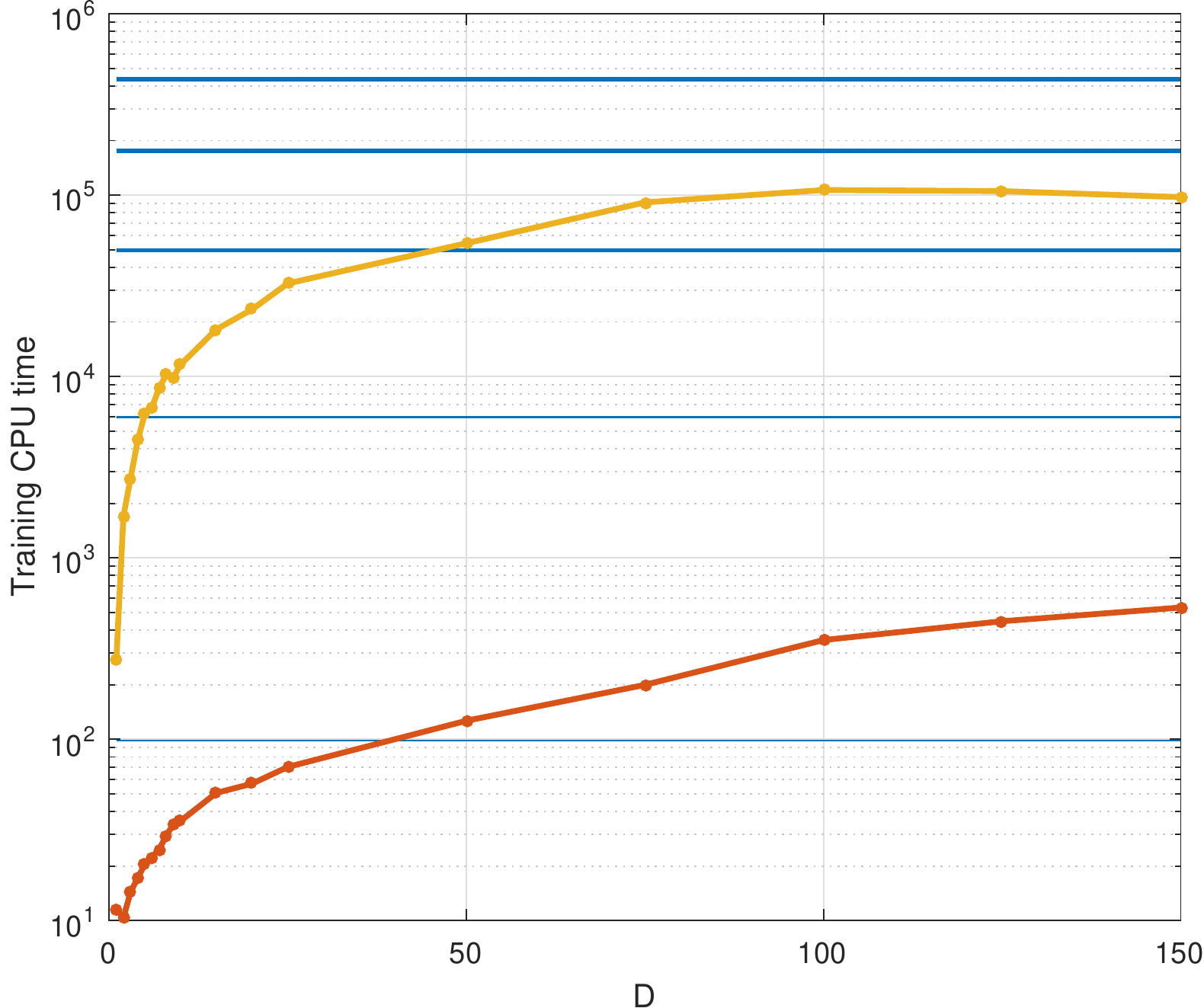}
\\
\includegraphics[width=.35\textwidth]{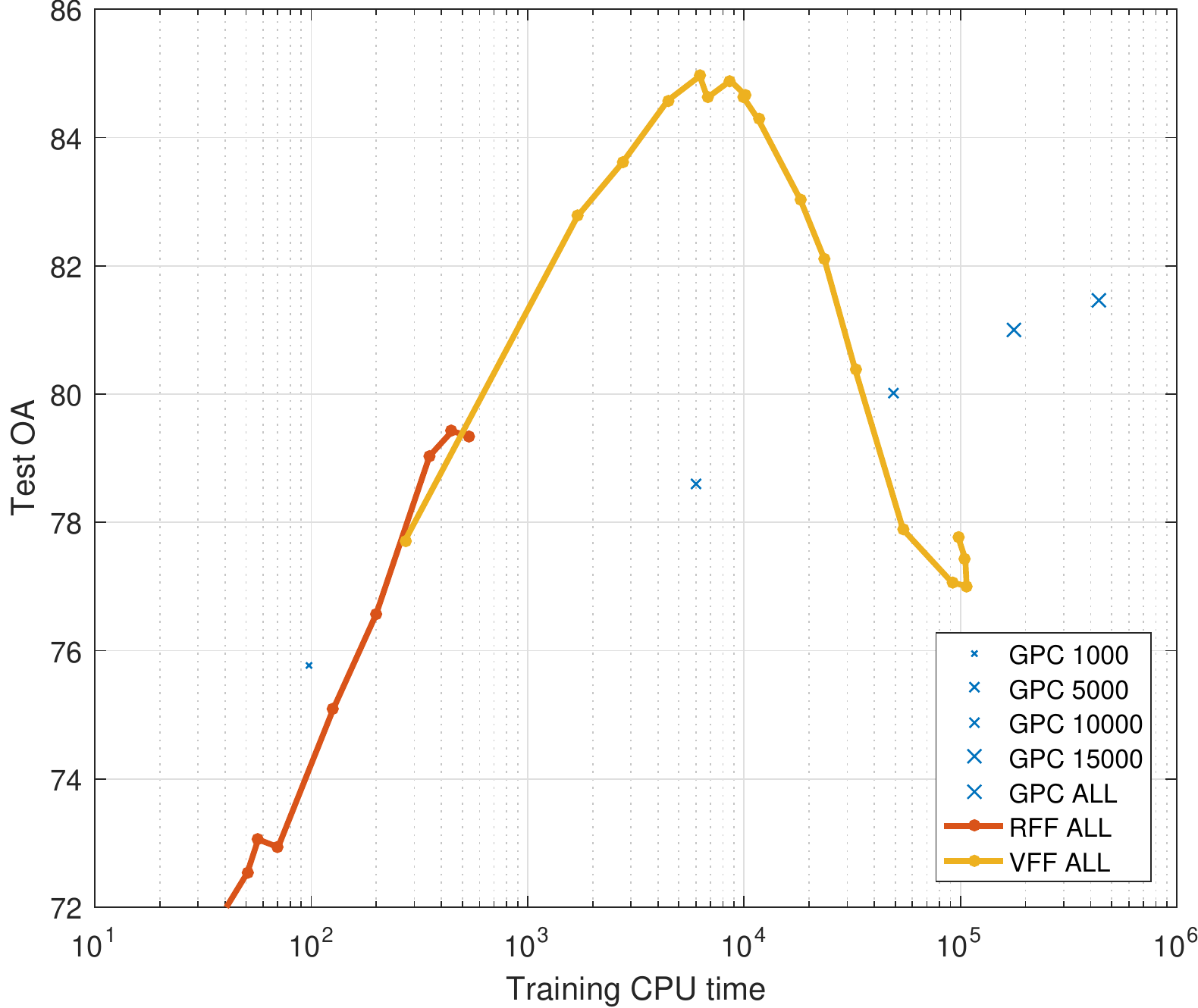}
&
\includegraphics[width=.35\textwidth]{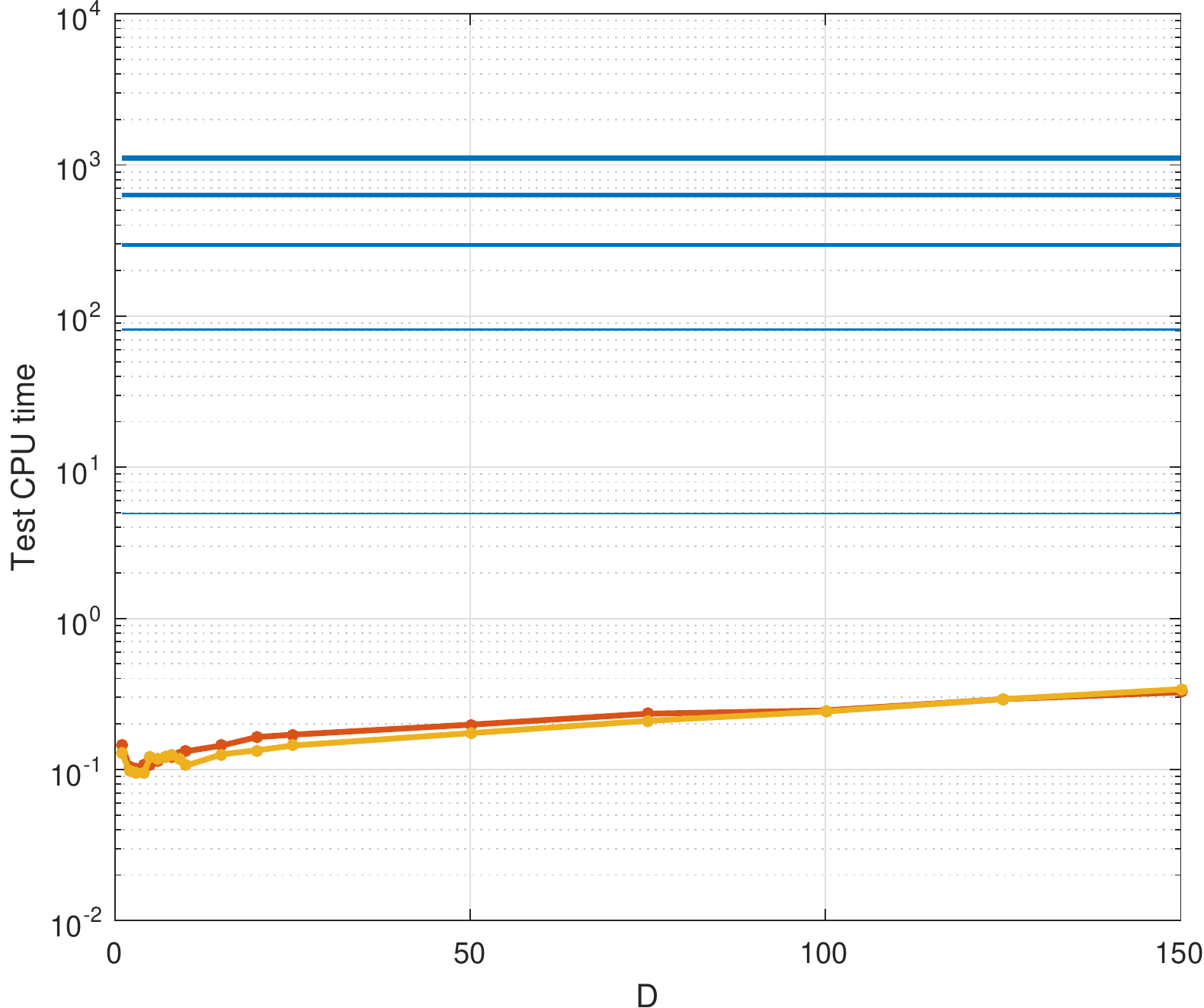}
\end{tabular}
\caption{
Experimental results for the IAVISA dataset. From left to right and top to bottom, the first plot shows the test overall accuracy (OA) of RFF-GPC, VFF-GPC, and GPC for the different values of $n$ (number of training examples) and $D$ (number of Fourier frequencies) considered. 
The second column is analogous, but displays the CPU time needed to train each method (instead of the test OA). 
The third column summarizes the two previous ones, providing a trade-off between test OA and training CPU time.
The last column is analogous to the first and second ones, but showing the CPU time used at the test step.
The legend for second and fourth plots is the same as the one in the first plot. However, in the third plot the GPC lines degenerate into single points (since GPC does not depend on $D$). In both legends, the numbers indicate the amount $n$ of training examples used, which determines the width/size of the lines/points too (\textrm{ALL} means the whole training dataset, i.e. $n\approx 20000$). As explained in the main text, the results are the mean over five independent runs.
}
\label{fig:IAVISA}
\end{center}
\end{figure*}

As explained in Section \ref{sec:data_IAVISA}, this problem involves a total amount of $n=24923$ instances.
We performed five-fold cross-validation, which produces five pairs of training/test datasets with (approximately) $20000/5000$ instances each. Results are then averaged. 
Since RFF-GPC and VFF-GPC are conceived for large-scale applications (they scale linearly with $n$, recall their $\mathcal{O}(nD^2)$ training cost), they will not be utilized with values of $n$ lower than this training dataset size of $\approx 20000$ (even GPC is able to cope with this size).
Indeed, in the case of GPC we use the values $n\in\{1000,5000,10000,15000, \mathrm{ALL}\approx20000\}$.
Regarding the number of Fourier frequencies $D$, we consider the grid $D\in\{1:10, 15:5:25, 50:25:150\}$. 
The experimental results, which include the same metrics as those used for the previous problem on landmarks, are shown in Figure \ref{fig:IAVISA}. 

In this case, we again observe a clear outperformance of VFF-GPC against GPC: it achieves higher test OA while requiring less training and test CPU times. Moreover, the improvement in test OA is greater than $3\%$, and train/test CPU times are around $100$ and $1000$ times lower respectively.
However, unlike in the previous problem of cloud detection over landmarks, RFF-GPC does not exhibit such a clear superiority over GPC in this application. Whereas it does drastically decrease the train/test CPU times, it is not able to reach the test OA of GPC with $n=10000$.
Therefore, in practice, the optimal choice for this application is VFF-GPC. RFF-GPC would only be recommended if the training CPU time is a very strong limitation.

The main reason why RFF-GPC is not completely competitive in this problem is its theoretical scope: as an efficient approximation to GPC, it is conceived for large scale applications which are out of the reach of standard GPC. If the size of the problem allows for using GPC (as in this case), then RFF-GPC will only provide a more efficient alternative (less training and test CPU times), but its predictive performance will be always below that of GPC. Moreover, the difference in this performance is directly influenced by the original dimension $d$ of data (recall that the kernel approximation behind RFF-GPC exponentially degrades with $d$, Section \ref{sec:theory_RFF}). This is precisely a second hurdle that RFF-GPC finds in IAVISA: the high $d=100$ makes RFF-GPC with the full dataset be quite far from the corresponding GPC at predictive performance (test OA). In conclusion, the ideal setting for RFF-GPC is a large scale problem (high $n$) with few features (low $d$), precisely the opposite to the IAVISA dataset.

Interestingly, VFF-GPC bypasses these limitations of RFF-GPC by {\em learning a new kernel and not just approximating} the SE one. First, VFF-GPC is not just a GP adaptation well-suited for large scale applications, but a general-purpose, expressive, and very competitive kernel-based classifier that scales well with the number of training instances. Second, as it does not rely on the kernel approximation, VFF-GPC is not affected by the original dimension $d$ of data. Both ideas are empirically supported by the results obtained in IAVISA.

\begin{figure}[htp!]
\begin{center}
\small
\includegraphics[width=.35\textwidth]{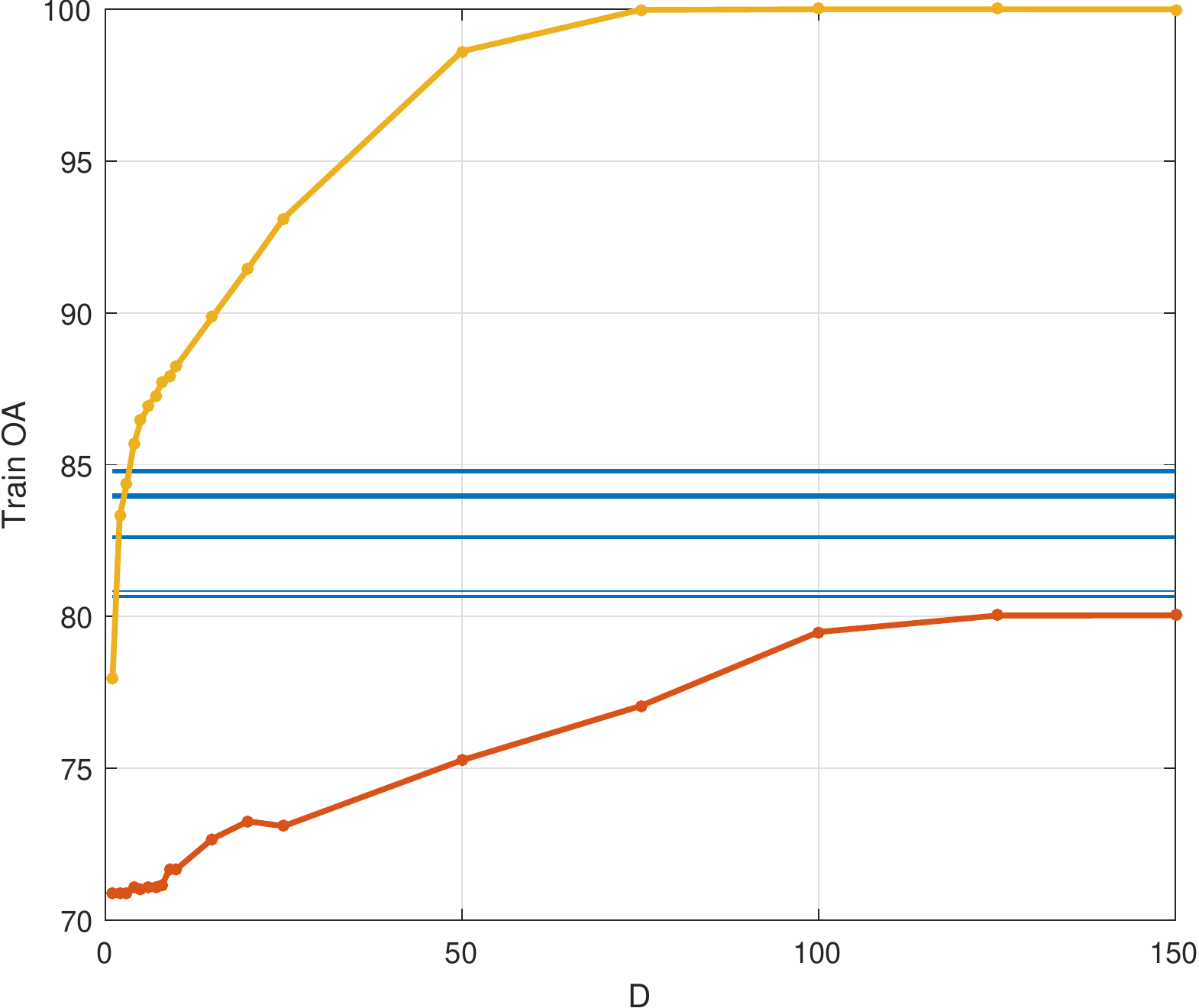}
\vspace{-0.25cm}
\caption{Train OA in the IAVISA dataset for RFF-GPC, VFF-GPC, and GPC with different values of $n$ (number of training examples) and $D$ (number of Fourier frequencies). These results complement the first plot in Figure \ref{fig:IAVISA}, showing that high values of $D$ make VFF-GPC over-fit to the training dataset. The legend and its interpretation are the same as there.}
\label{fig:IAVISA2}
\end{center}
\end{figure}

The first plot of Figure \ref{fig:IAVISA} shows that the predictive performance of VFF-GPC does not necessarily improves by increasing $D$.
This is the expected behavior from the theoretical formulation of VFF-GPC, where the Fourier frequencies are $D$ parameters to be estimated.
Thus, a higher amount of them confers VFF-GPC a greater flexibility to learn hidden patterns in the training dataset, but also the possibility to over-fit very particular structures of it which do not generalize to the test set.
This is the classical problem of the \emph{model complexity} in machine learning, and it is further illustrated in Figure \ref{fig:IAVISA2}. Together with the first plot in Figure \ref{fig:IAVISA}, it shows the paradigmatic behaviour of train and test performance in presence of over-fitting: train OA grows with the model complexity (great flexibility allows for learning very particular structures of the training set, even reaching a $100\%$ of train OA), whereas test OA initially grows (the first patterns are part of the ground truth and thus general to the test set) but then goes down (when the learned information is too specific to the training set).
Notice that this over-fitting phenomena did not occur at LANDMARKS, where test OA monotonically increased with $D$. In addition to the different nature of the problems, the training dataset size $n$ plays a crucial role at this: smaller datasets (like IAVISA) are more prone to over-fitting than larger ones (LANDMARKS) under the same model complexity.

Finally, it is worth noting that VFF-GPC achieves its maximum test OA when using just $D=5$ Fourier frequencies. This reflects (i) a not very sophisticated internal structure of the IAVISA dataset (since just $5$ directions are enough to correctly classify $85\%$ of the data), and (ii) the VFF-GPC capability to learn those discriminative directions from data.
In particular, this shows that VFF-GPC can be used not only as a classifier, but also as a method that learns the most relevant discriminative directions in a dataset.
Unfortunately, RFF-GPC is not able to benefit from these privileged directions that may exist in some datasets, since it randomly samples and fix the Fourier frequencies from the beginning.


\subsection{Explicit classification maps for cloud detection}

The last two sections were dedicated to thoroughly analyze the performance of the proposed methods, empirically understand their behavior, weaknesses, and strengths, and compare them against GPC. In order to illustrate the explicit cloud detection behind the experiments, here we provide several explanatory classification maps obtained by the best model (in terms of predictive performance) for the LANDMARKS dataset: VFF-GPC with $n=300000$ and $D=200$. 

The classification maps are obtained for the whole year 2010 at the Dakhla landmark, 
with a total of 34940 satellite acquisitions. The acquired window size is $26\times20$ pixels. Relying on the proposed feature extraction procedure, we trained the four necessary models (high, mid, low, night), and then proceed to predict over the whole available amount of chips acquired in the 2010 year.\footnote{A full video with all the classification maps is available at \url{http://decsai.ugr.es/vip/software.html} and \url{http://isp.uv.es/code/vff.html}. RFF-GPC and VFF-GPC codes are also provided.}

\begin{figure}[ht!]
\begin{center}
\includegraphics[width=.24\columnwidth]{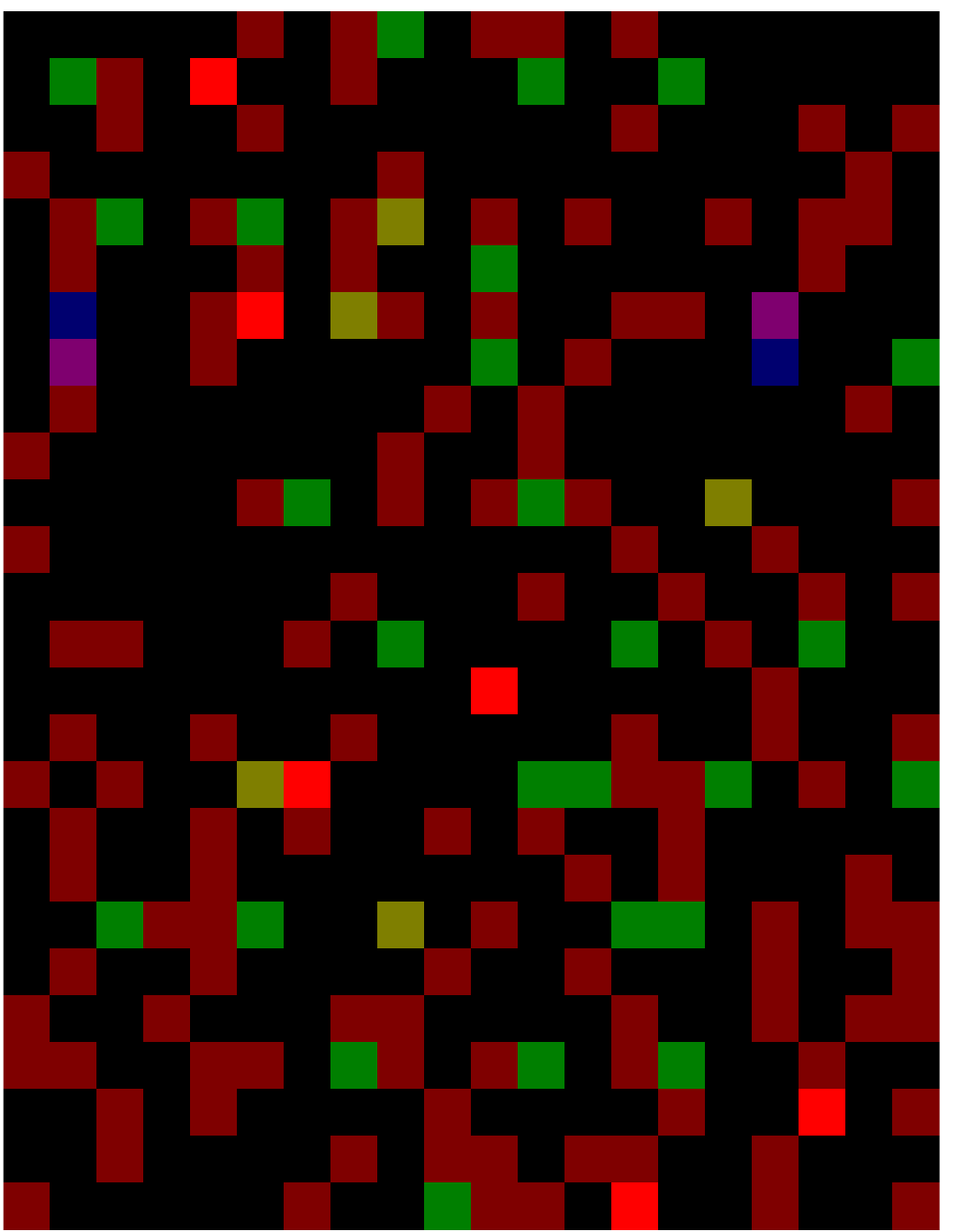}
\includegraphics[width=.24\columnwidth]{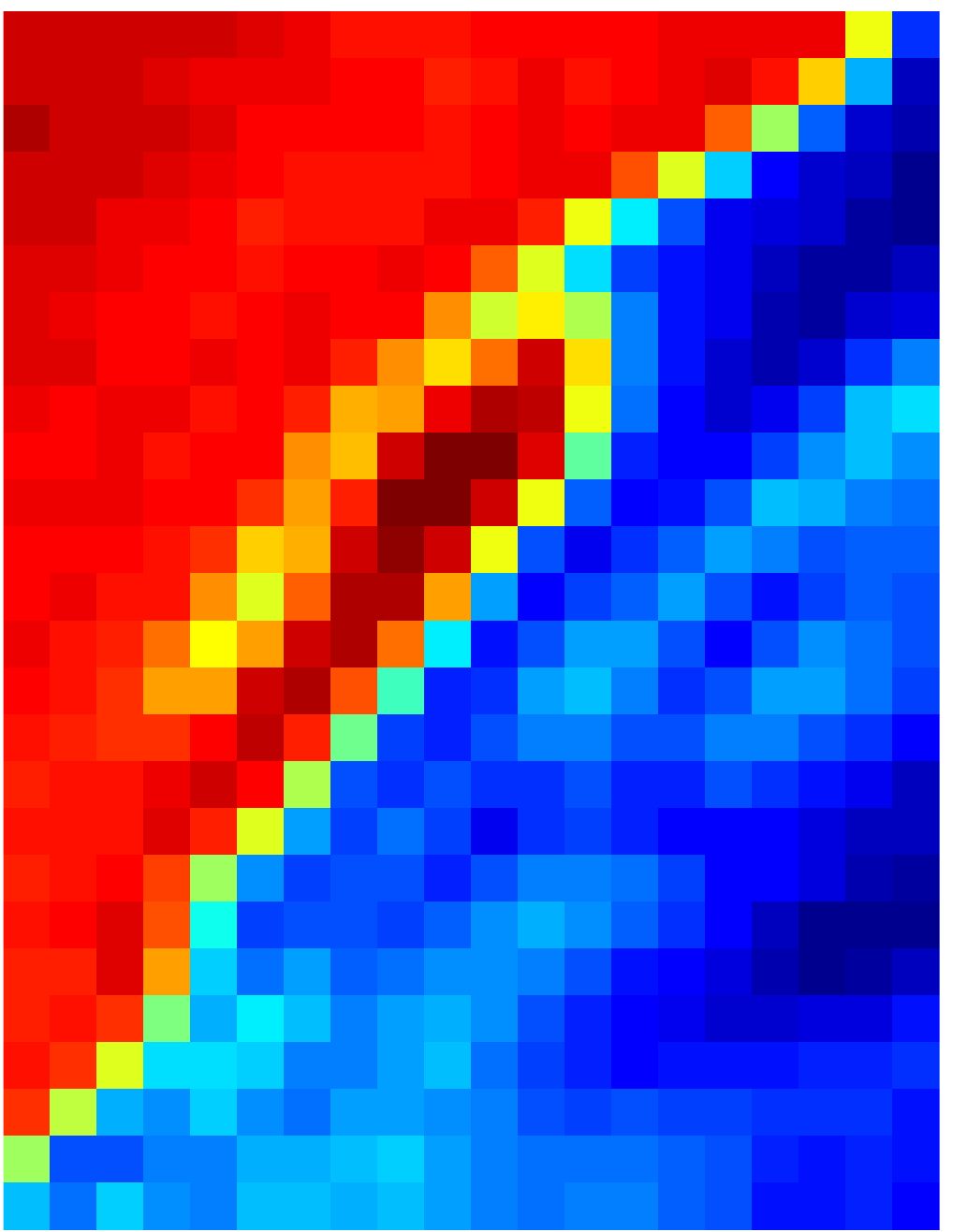}
\includegraphics[width=.24\columnwidth]{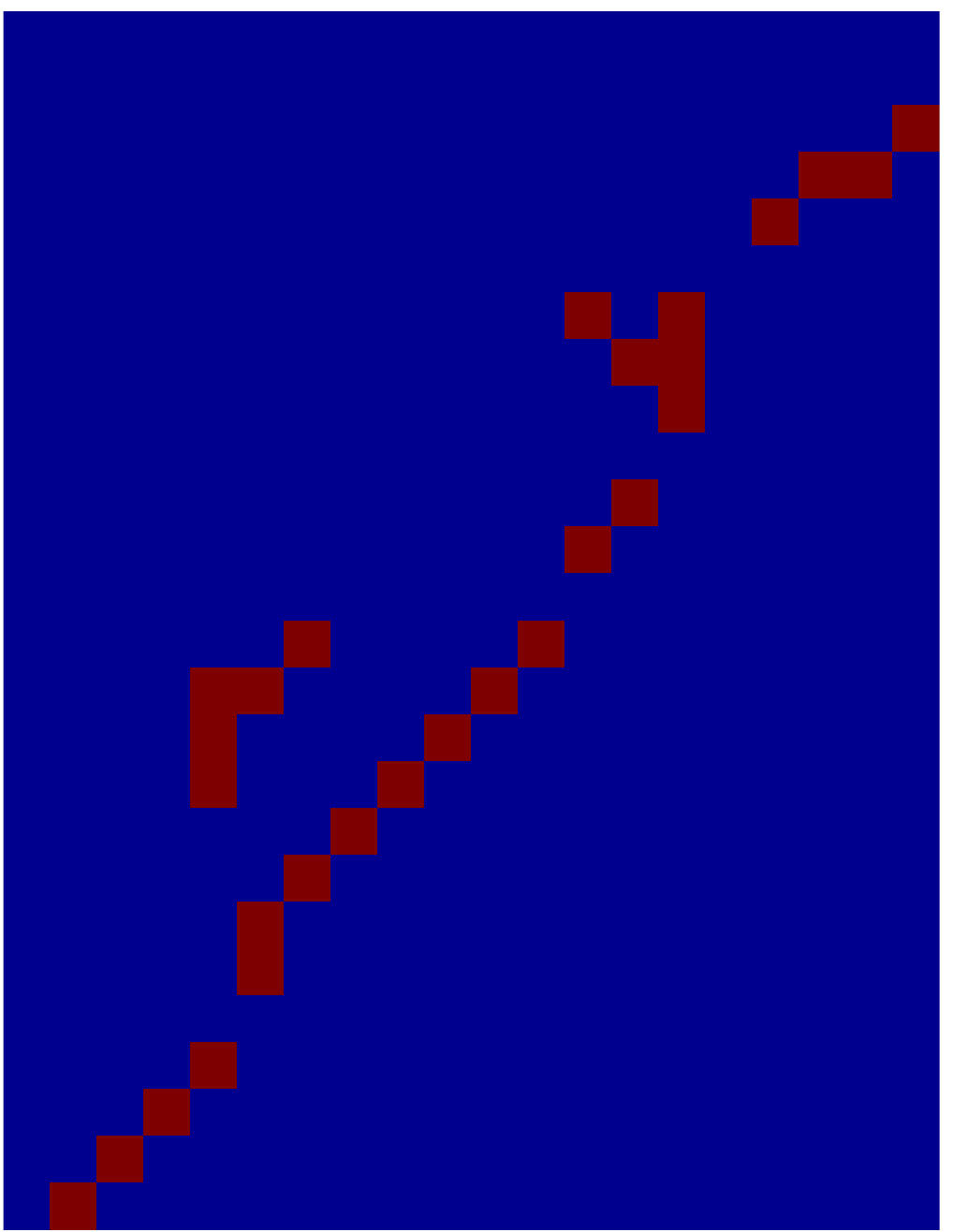}
\includegraphics[width=.24\columnwidth]{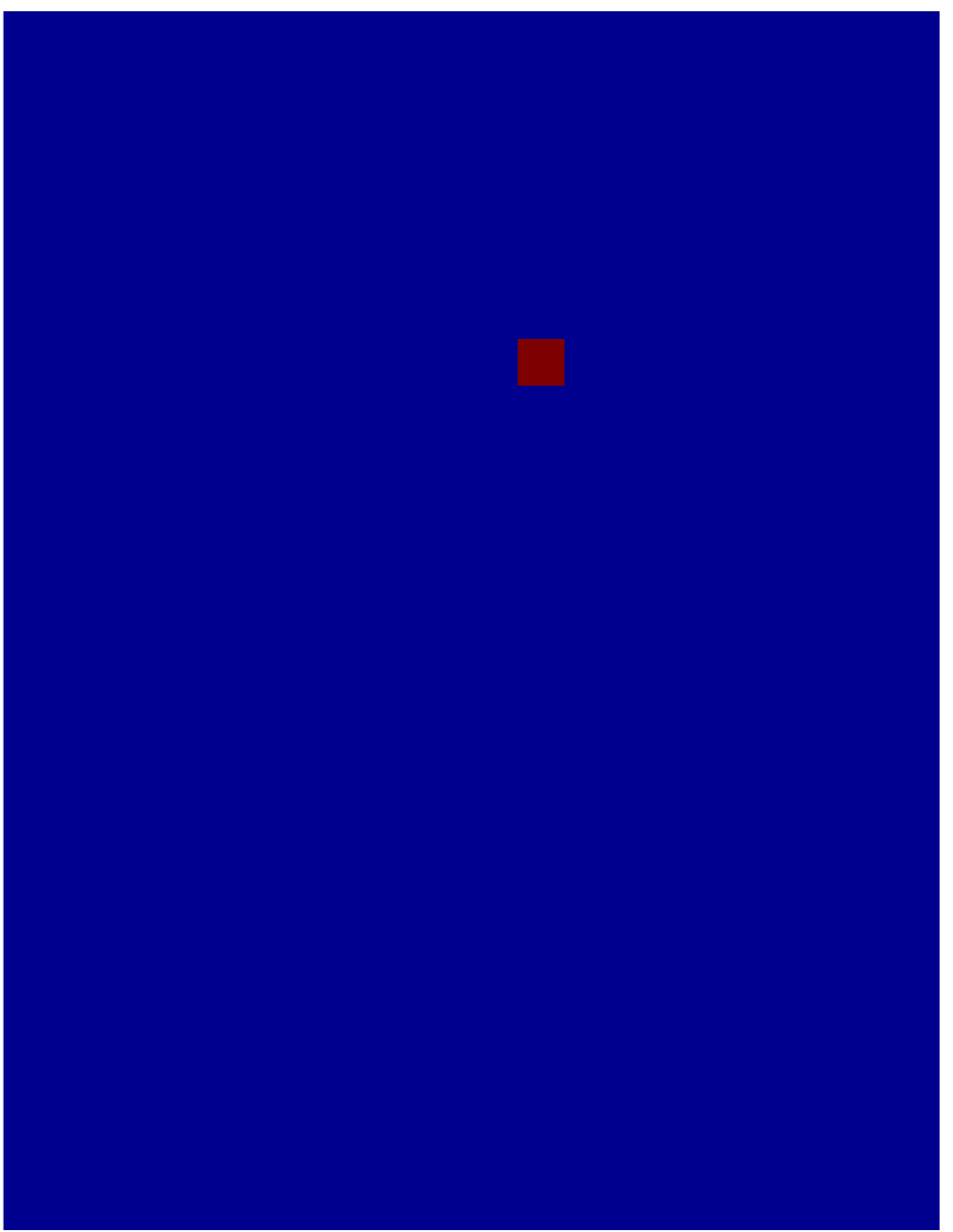}
\\
\includegraphics[width=.24\columnwidth]{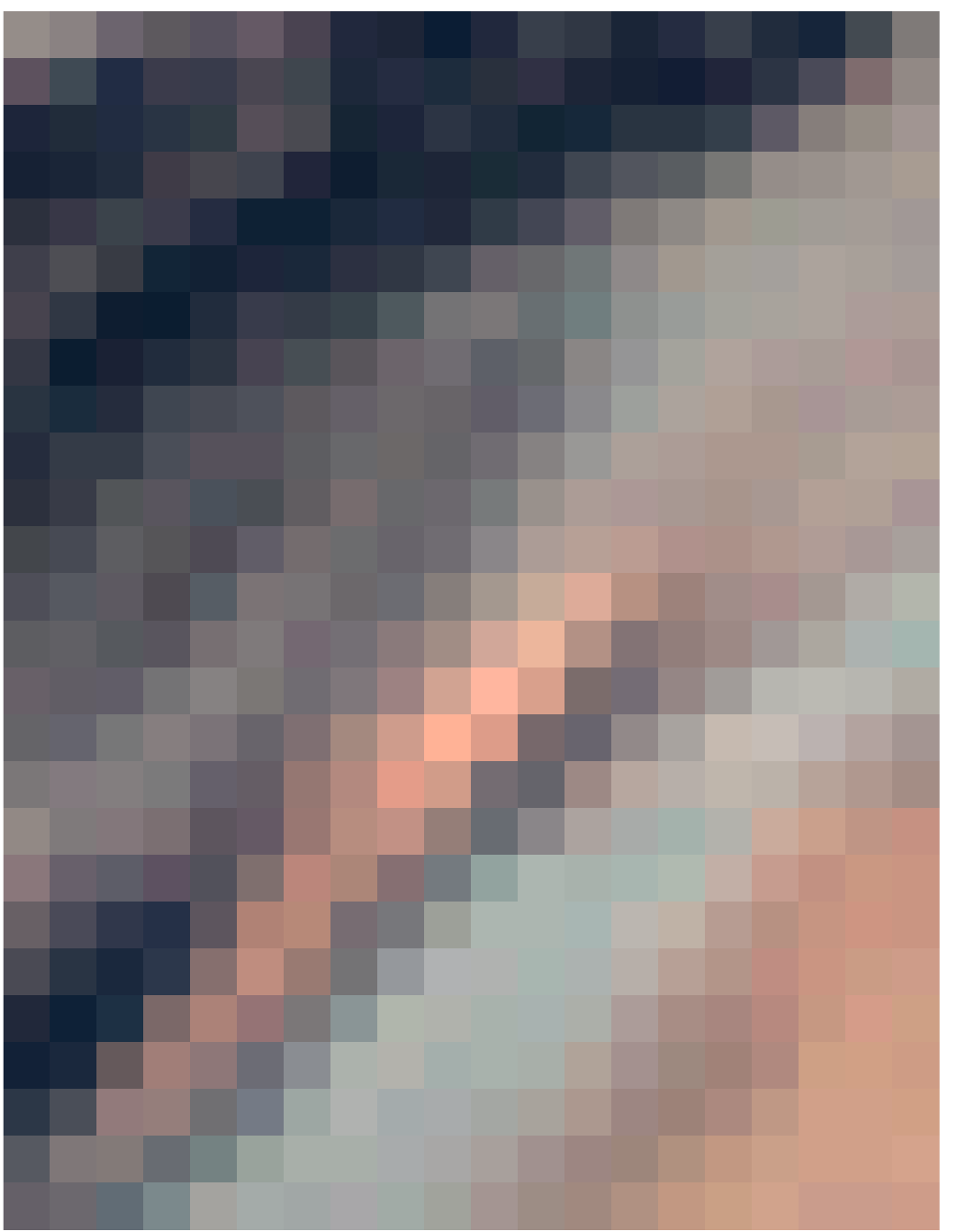}
\includegraphics[width=.24\columnwidth]{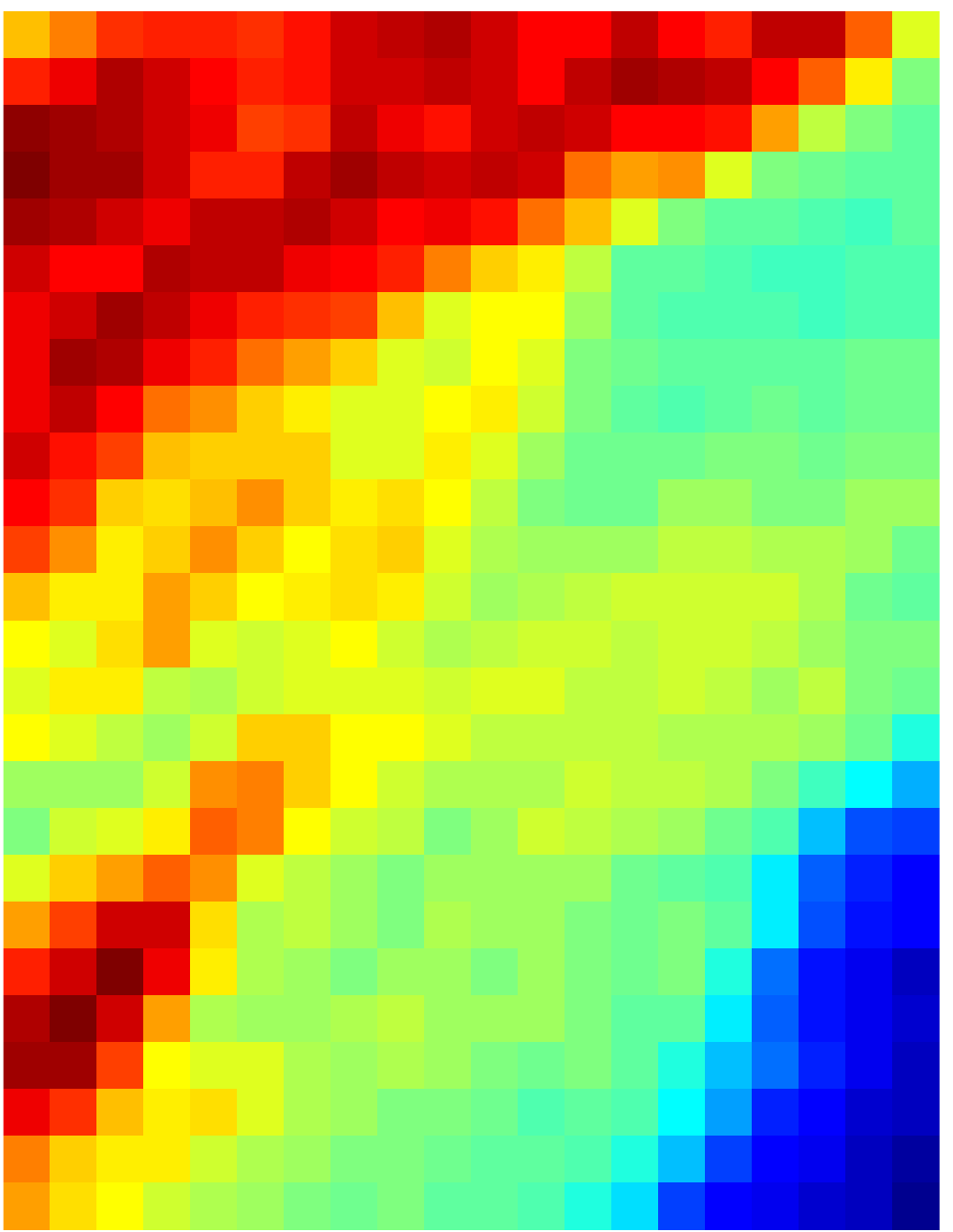}
\includegraphics[width=.24\columnwidth]{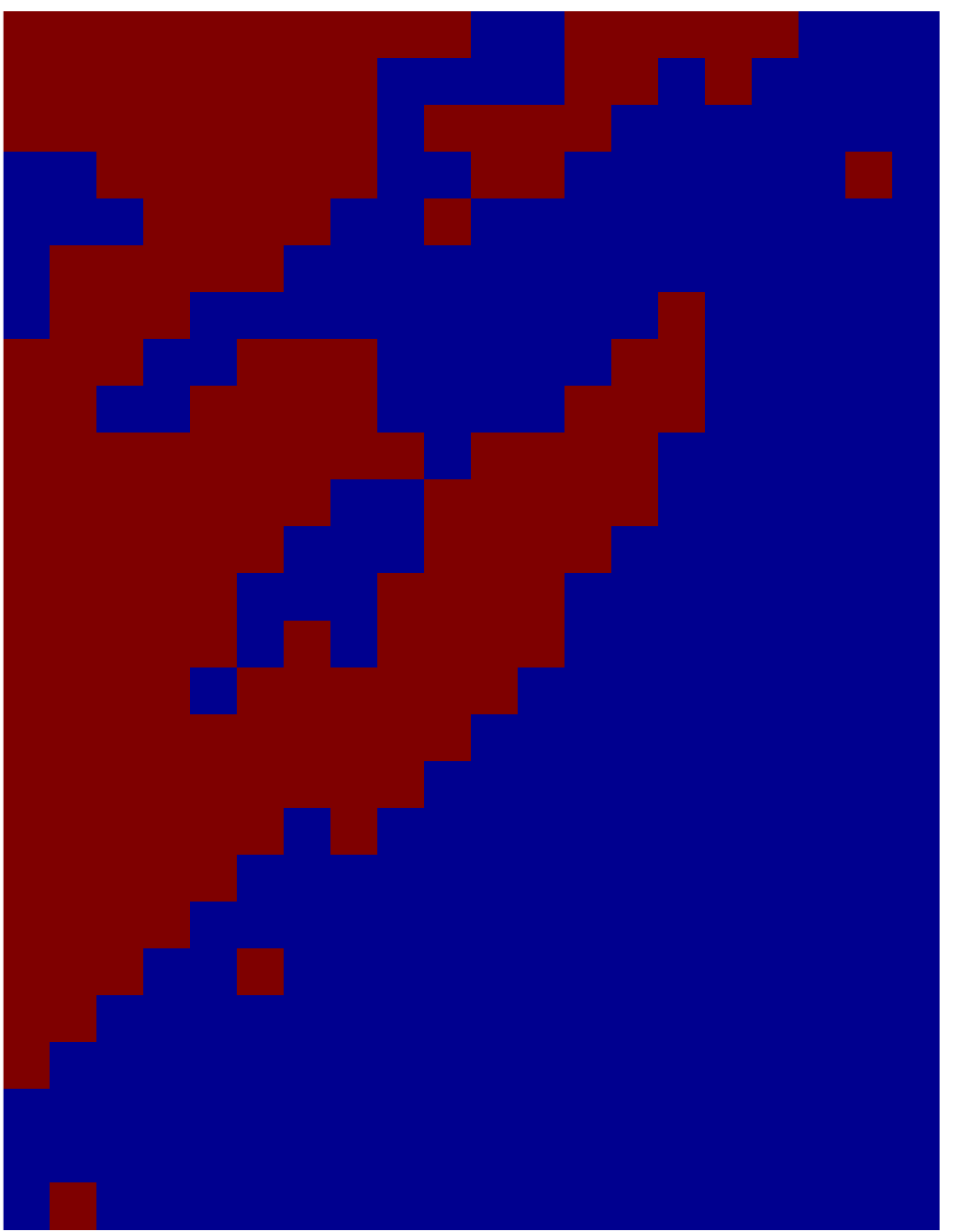}
\includegraphics[width=.24\columnwidth]{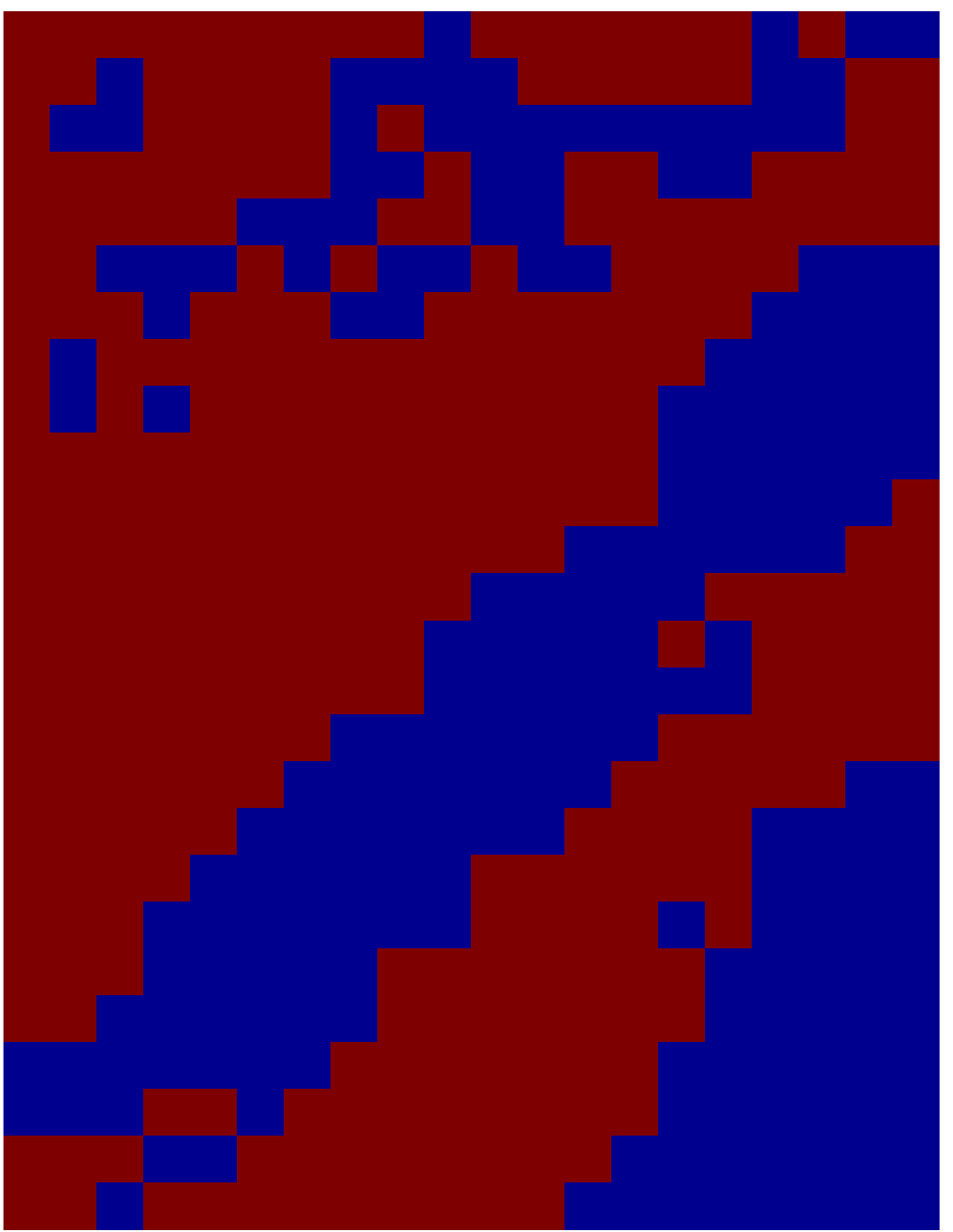}
\\
\includegraphics[width=.24\columnwidth]{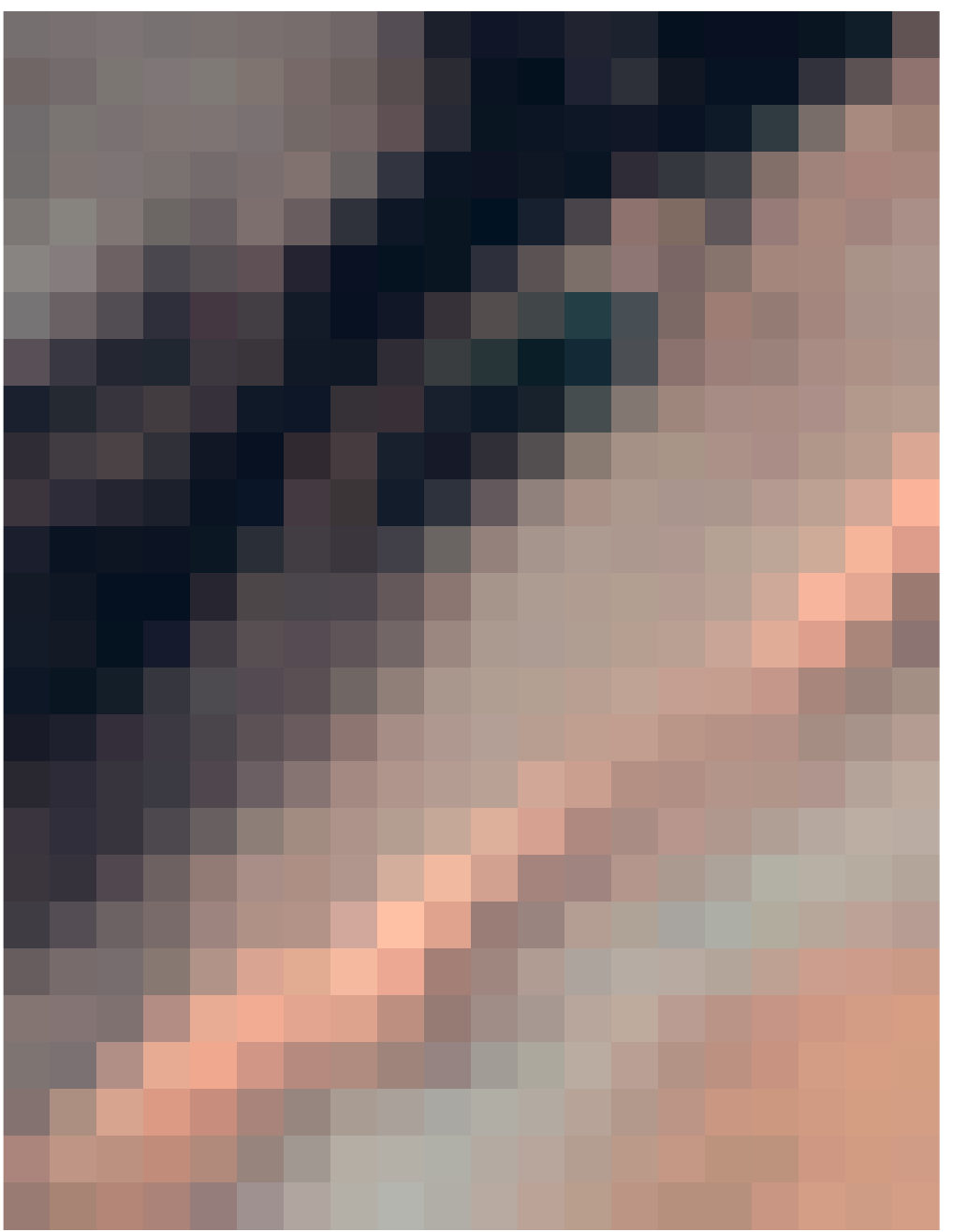}
\includegraphics[width=.24\columnwidth]{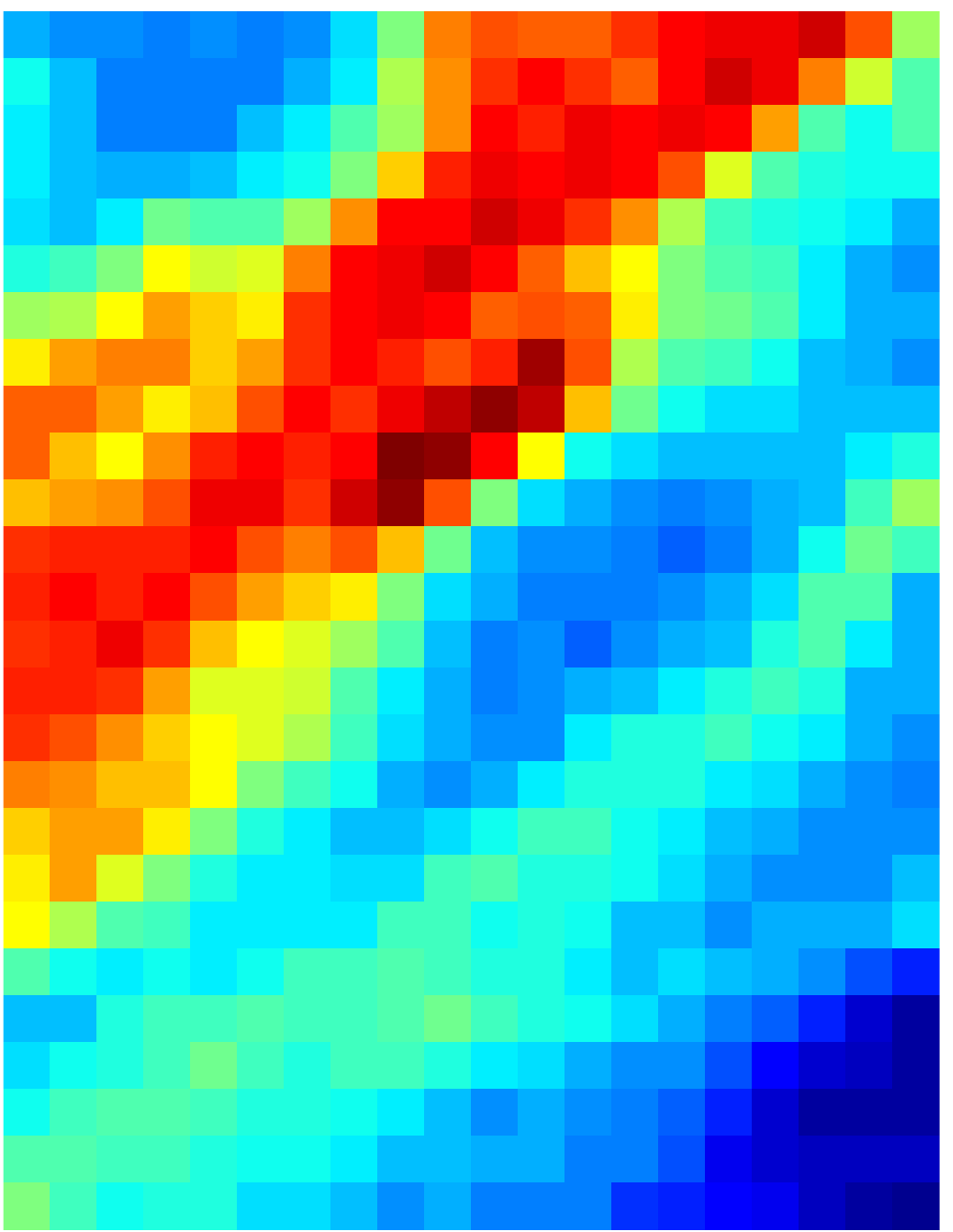}
\includegraphics[width=.24\columnwidth]{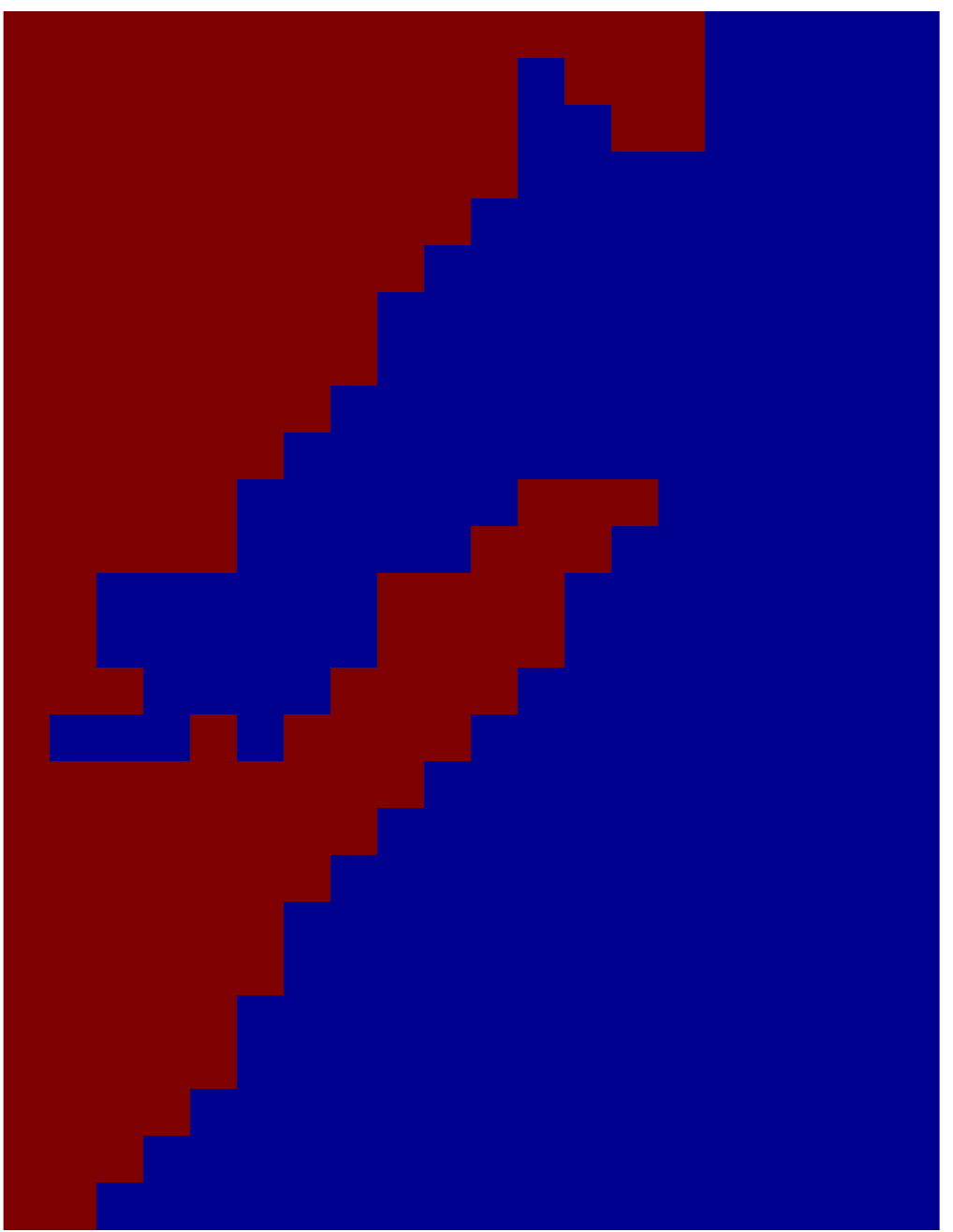}
\includegraphics[width=.24\columnwidth]{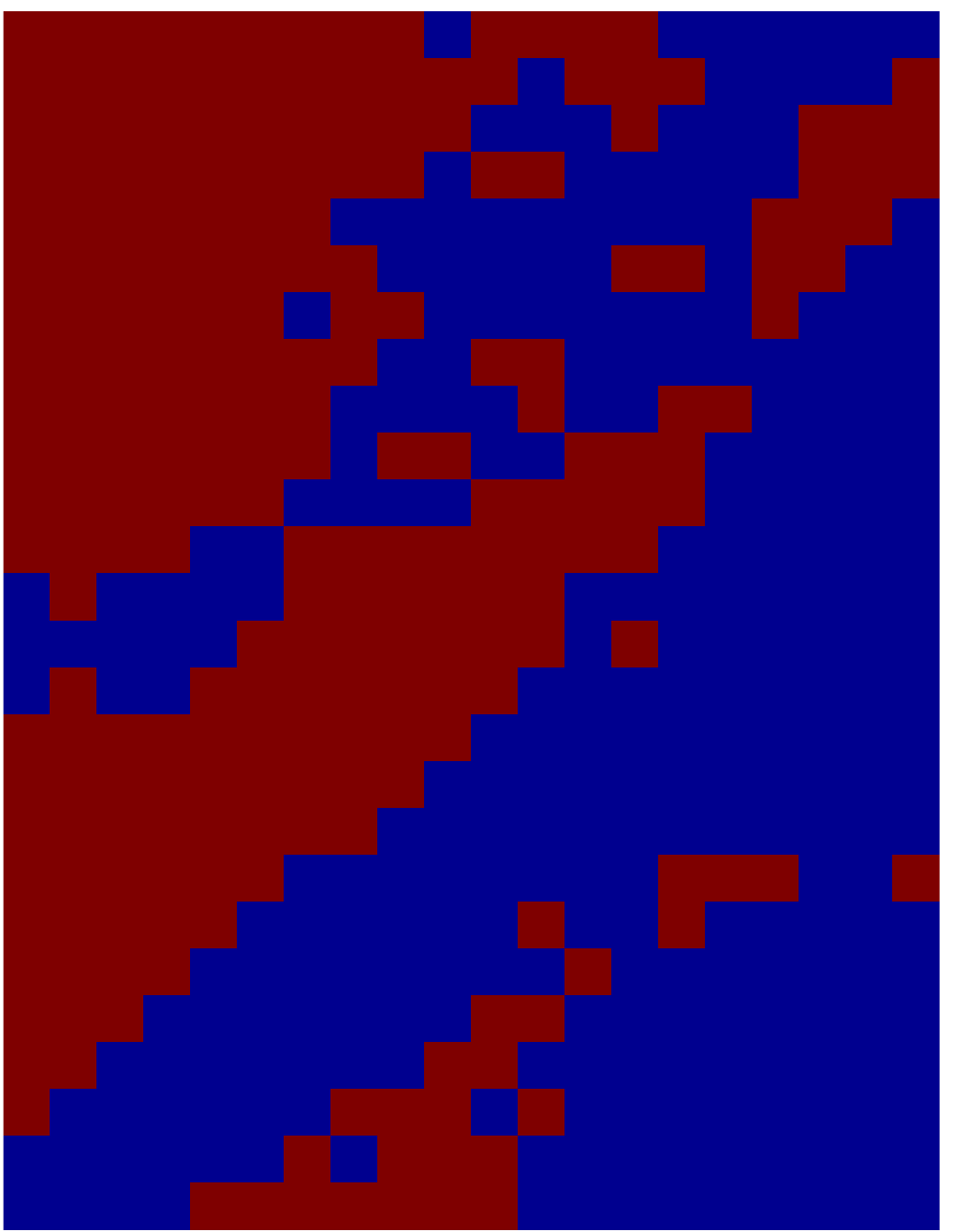}
\\
\includegraphics[width=.24\columnwidth]{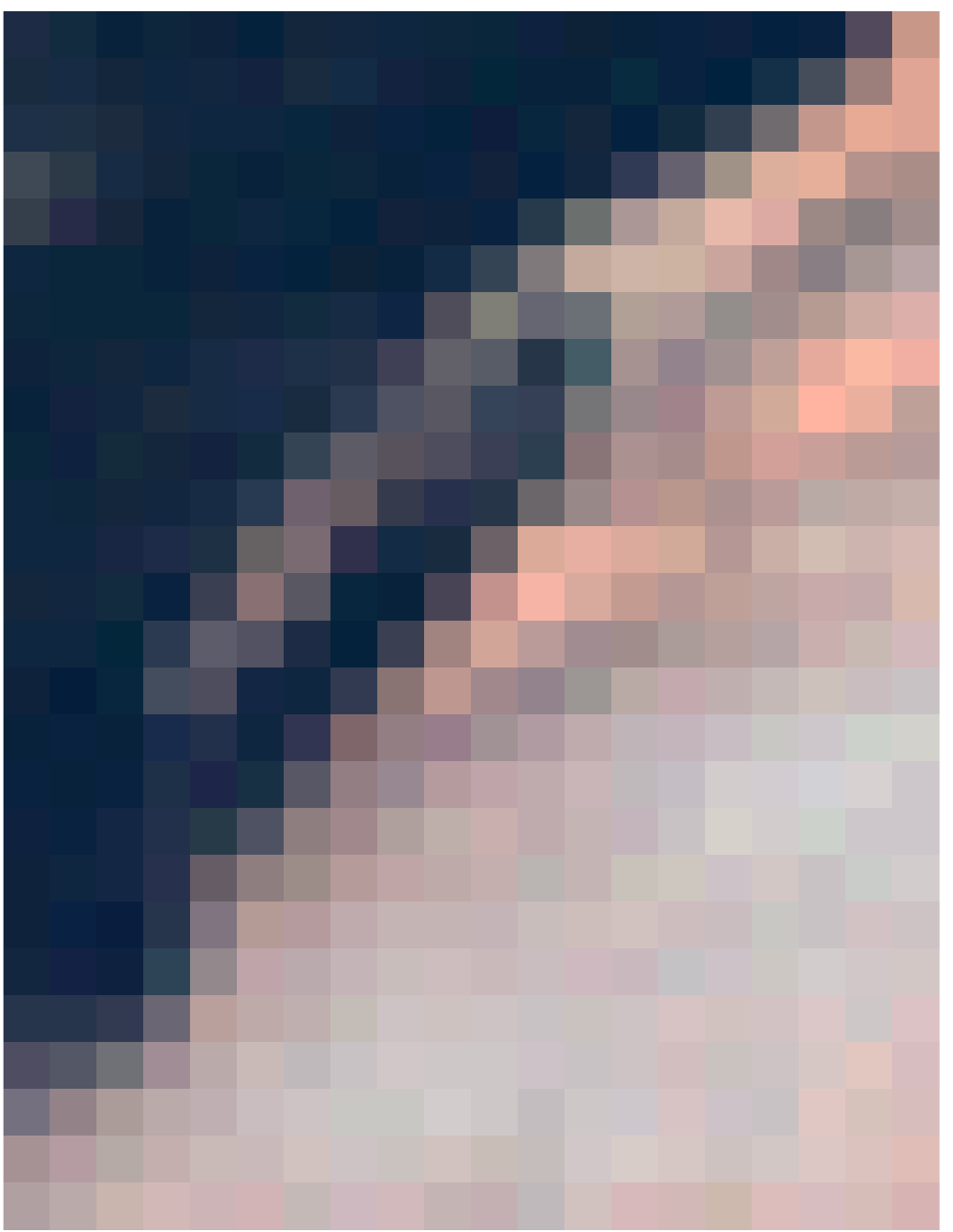}
\includegraphics[width=.24\columnwidth]{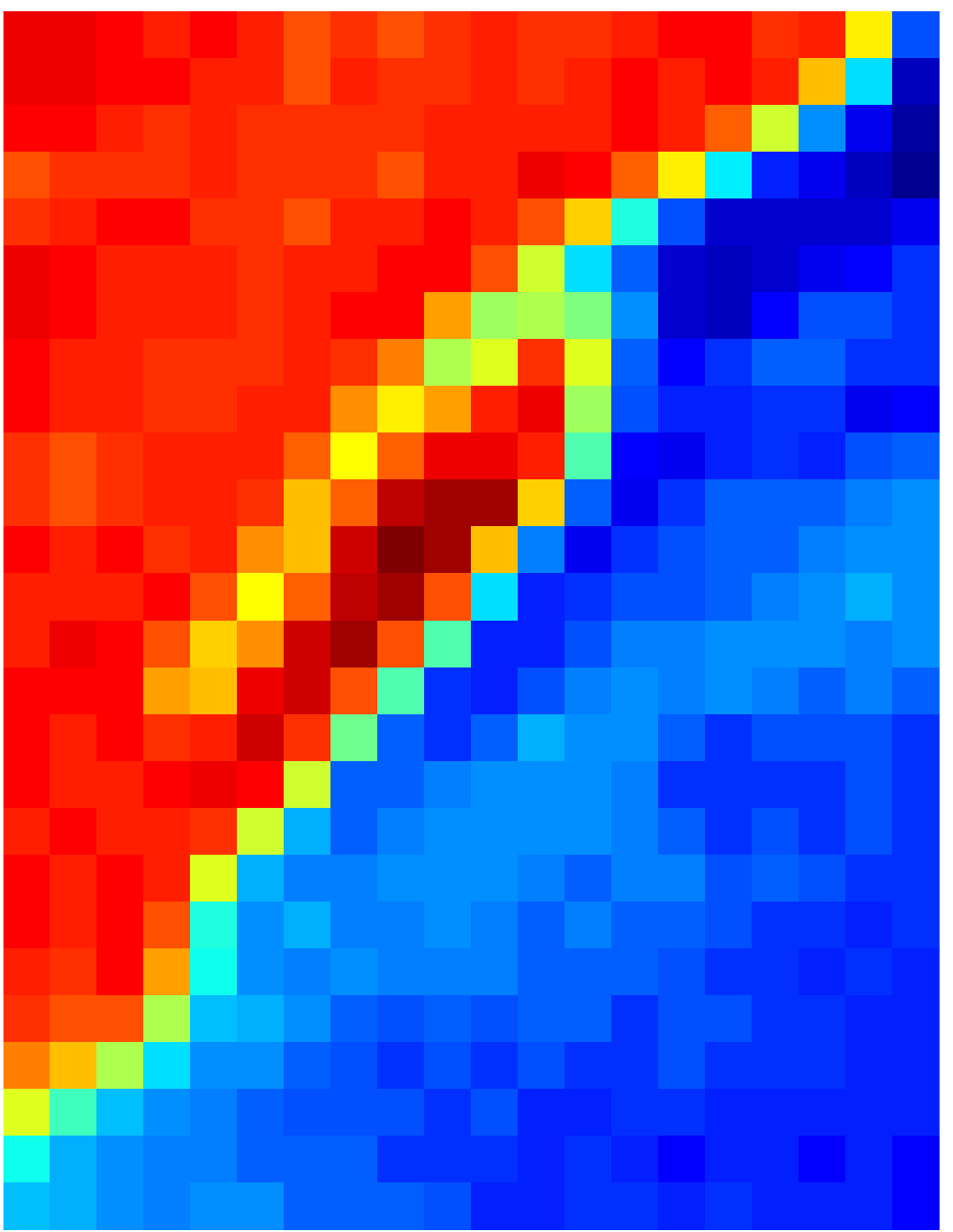}
\includegraphics[width=.24\columnwidth]{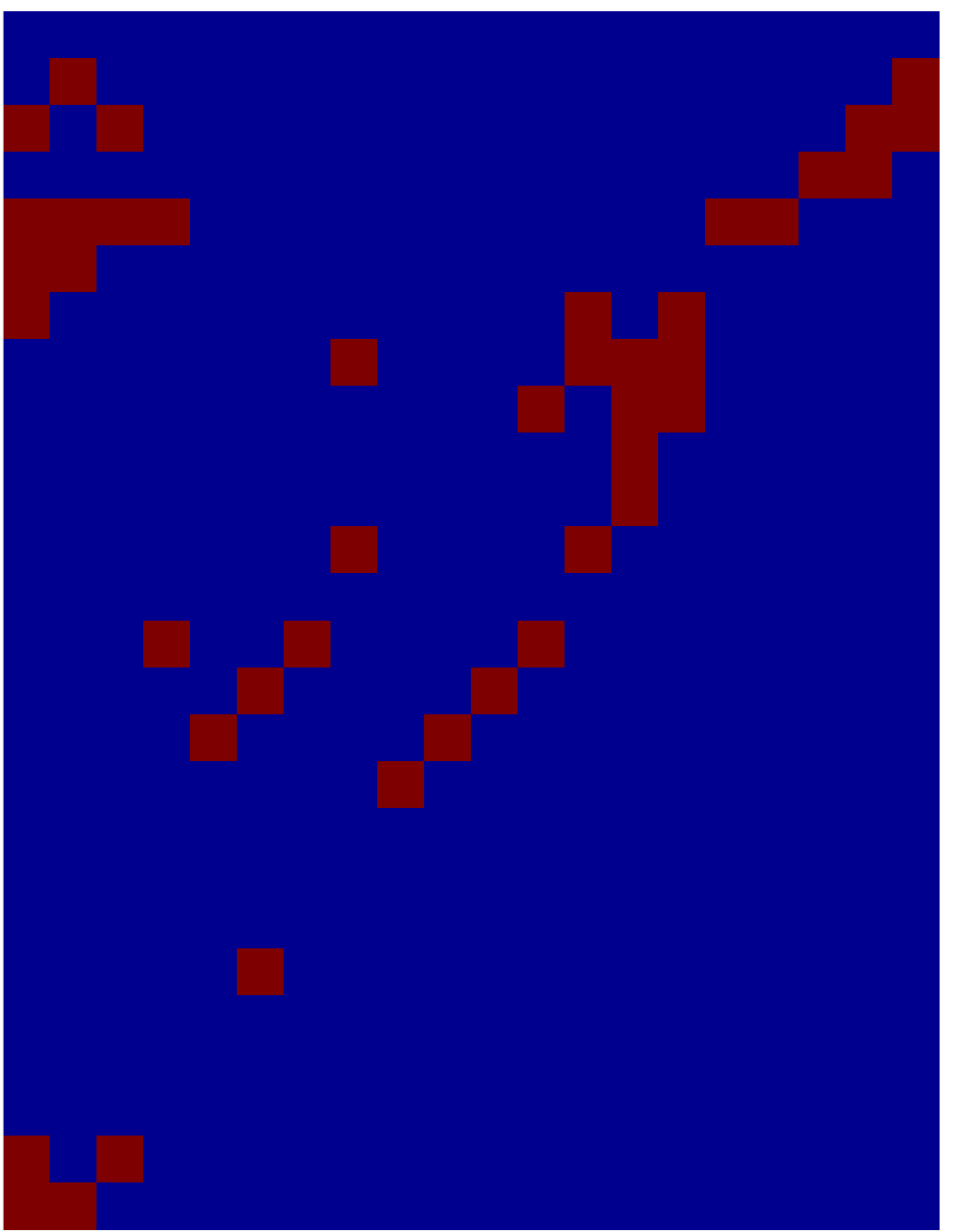}
\includegraphics[width=.24\columnwidth]{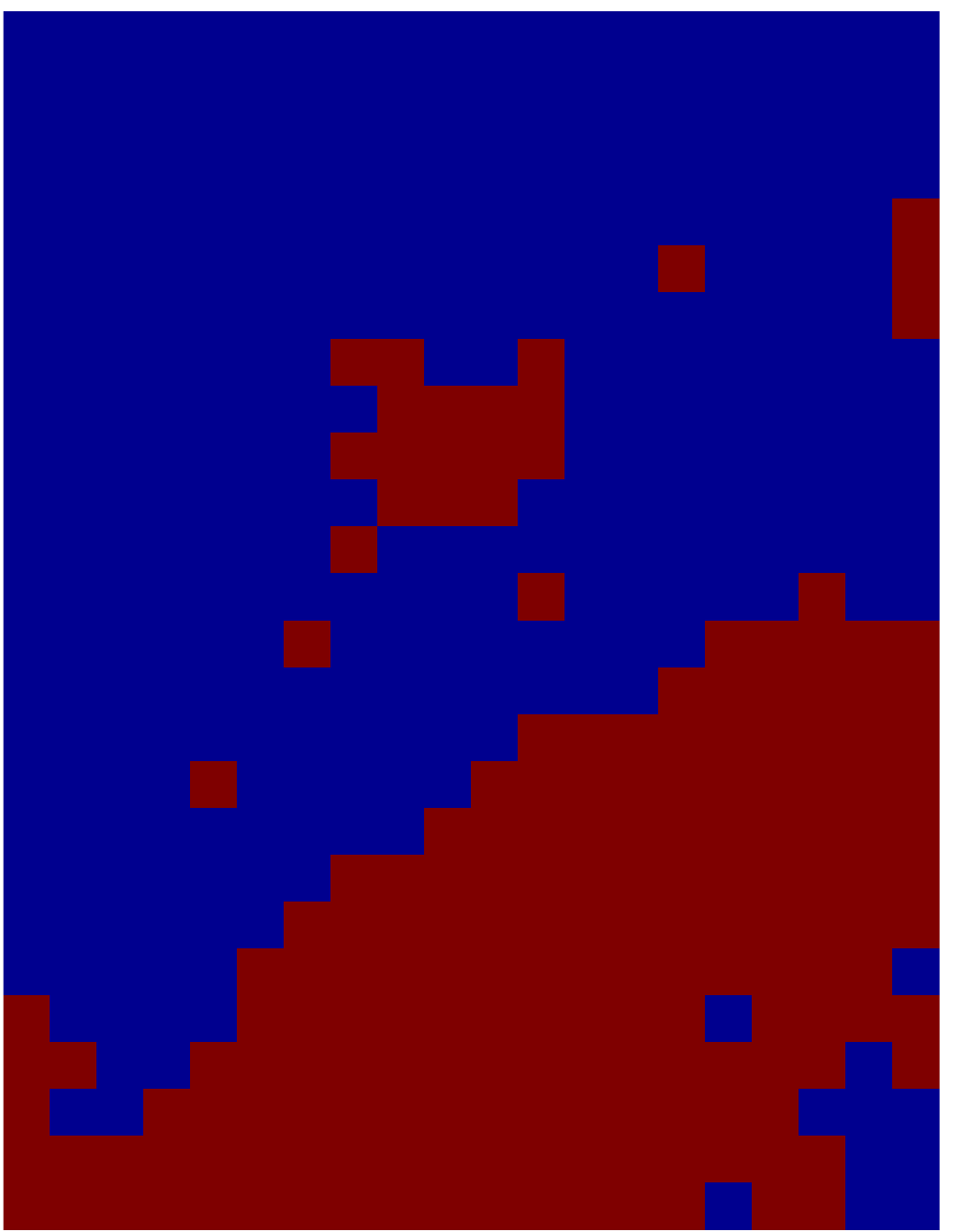}

\vspace{-0.25cm}
\caption{Explicit classification maps for the Dakhla landmark. The rows correspond with four different acquisitions.
The first column shows the visible RGB channels (which are not informative for \emph{night} acquisitions such us the first one), the second column is the infrared 10.8$\mu m$ spectral band (very illustrative in night scenarios), the third column represents the ground truth obtained by EUMETSAT (the L2 cloud mask), and the last one is the VFF-GPC classification map. In the last two columns, the red color is used for cloudy pixels and blue for cloud-free ones.}
\label{fig:classification-maps}
\end{center}
\end{figure}

In Figure~\ref{fig:classification-maps}, several chips are provided with the aim of illustrating different behaviors.
In the first situation (first row), we can see a characteristic error of the L2 cloud mask, which sometimes tends to wrongly label the coastline pixels as cloudy. However, VFF-GPC leads to a better classification, identifying just one cloudy pixel and thus avoiding this negative coastline effect\footnote{As a clarification note, the coastline pixels were removed from the training dataset by applying a carefully designed morphological filter around coastlines.}.
In the second row, the visual channels show large clouds crossing the landmark. In the bottom-right of the image, a long cloud is unlabeled in the L2 cloud mask but correctly detected by VFF-GPC. While being formally accounted as an error, such discrepancy is actually positive for our method. Moreover, VFF-GPC shows an interesting cloud-sensitive behaviour at the top-left cloudy mass, identifying a larger cloudy area than that provided by EUMETSAT. This is a desirable propensity in cloud detection applications, where we prefer to identify larger clouds (and then thoroughly analyze them) rather than missing some of them. 
In the third row, the RGB channel allows for visually identifying three main cloudy masses at the landmark.
The L2 mask poorly labels the central cloudy band, and does not detect the lower cloud. Both deficiencies are overcome by VFF-GPC.
Finally, the fourth chip shows a huge cloudy mass that is undetected by the L2 mask but is correctly identified by VFF-GPC. 

Therefore, although VFF-GPC is trained with an imperfect ground truth, we observe that it is able to bypass some of these deficiencies, and exhibits a desirable cloud-sensitive behavior. This improvement can be also related to the particular design of the training datasets, splitting the problem into four different cases depending on the illumination conditions.

\section{Conclusions and Future Work}\label{sec:conclusions}

We presented two efficient approximations to Gaussian process classification to cope with big data classification problems in EO. The first one, RFF-GPC, performs standard GP classification by means of a fast approximation to the kernel (covariance) via $D$ random Fourier features. The advantage of the method is mainly computational, as the  training cost is ${\mathcal O}(nD^2)$ instead of the ${\mathcal O}(n^3)$ induced by the direct inversion of the $n\times n$ kernel matrix (the test cost is also reduced to be independent on $n$, from $\mathcal{O}(n^2)$ to $\mathcal{O}(D^2)$). The RFF method approximates the squared exponential (SE) covariance with Fourier features randomly sampled in the whole spectral domain. The solid theoretical grounds and good empirical performance makes it a very useful method to tackle large scale problems. Actually, the use of RFF has been exploited before in other settings, from classification with SVMs to regression with the KRR. However, we emphasize two main shortcomings. Firstly, the RFF approach can only approximate (theoretically and in practice) a predefined kernel (the SE one in this work). Secondly, by sampling the Fourier domain from a Gaussian, one has no control about the expressive power of the representation since some frequency components of the signal can be better represented than others. As a consequence, the approximated kernel may not have good discrimination capabilities. Noting these two problems, we proposed here our second methodology: a variational GP classifier (VFF-GPC) which goes one step beyond by optimizing over the Fourier frequencies. It is shown to be not just a GP adaptation well-suited for large scale applications, but a whole novel, general-purpose, and very competitive kernel-based classifier that scales well (linearly, as RFF-GPC) with the number of training instances.

We illustrated the performance of the algorithms in two real remote sensing problems of large and medium size. In the first case study, a challenging problem dealt with the identification of clouds over landmarks using Seviri/MSG imagery. The problem involved several hundred thousands data points for training the classifiers. In the second case study, we used the IAVISA dataset, which exploits IASI/AVHRR data to identify clouds with the IASI infrared sounding data. Compared to the original GPC, the experimental results show a high competitiveness in accuracy, a remarkable decrease in computational cost, and an excellent trade-off between both.

These results encourage us to expand the experimentation to additional problems, trying to exploit the demonstrated potential of VFF-GPC when dealing with any value of $n$ (training data set size) and $d$ (original dimension of the data). Other prior distributions and inference methods, as explained at the end of Section \ref{sec:theory_RFF},  will be also explored in the future.

\ifCLASSOPTIONcaptionsoff
  \newpage
\fi

\bibliographystyle{IEEEtran}
\bibliography{mybib}

\enlargethispage{-9.5cm}

\end{document}